\documentclass[10pt,twocolumn,letterpaper]{article}

\usepackage{cvpr}
\usepackage{times}
\usepackage{epsfig}
\usepackage{graphicx}
\usepackage{amsmath}
\usepackage{amssymb}

\usepackage{color}

\usepackage{multirow}
\usepackage{booktabs}

\usepackage[breaklinks=true,bookmarks=false]{hyperref}

\cvprfinalcopy %

\def\cvprPaperID{4016} %

\ifcvprfinal\fi
\begin{document}

\title{Rethinking the Evaluation of Video Summaries}

\author{Mayu Otani\\
{CyberAgent, Inc.}\\
\and
Yuta Nakashima\\
Osaka University\\
\and
Esa Rahtu\\
Tampere University\\
\and
Janne Heikkil\"{a}\\
University of Oulu\\
}

\maketitle

\begin{abstract}
Video summarization is a technique to create a short skim of the original video while preserving the main stories/content. There exists a substantial interest in automatizing this process due to the rapid growth of the available material. The recent progress has been facilitated by public benchmark datasets, which enable easy and fair comparison of methods. Currently the established evaluation protocol is to compare the generated summary with respect to a set of reference summaries provided by the dataset. In this paper, we will provide in-depth assessment of this pipeline using two popular benchmark datasets. Surprisingly, we observe that randomly generated summaries achieve comparable or better performance to the state-of-the-art. In some cases, the random summaries outperform even the human generated summaries in leave-one-out experiments. Moreover, it turns out that the video segmentation, which is often considered as a fixed pre-processing method, has the most significant impact on the performance measure. Based on our observations, we propose alternative approaches for assessing the importance scores as well as an intuitive visualization of correlation between the estimated scoring and human annotations.
\end{abstract}

\section{Introduction}

The tremendous growth of the available video material has escalated the demand for techniques that enable users to quickly browse and watch videos. One remedy is provided by the automatic video summarization, where the aim is to produce a short video skim that preserve the most important content of the original video. For instance, the original footage from a sport event could be compressed into a few minute summary illustrating the most important events such as goals, penalty kicks, etc.

\begin{figure}[t!]
    \centering
    \includegraphics[clip, width=\linewidth]{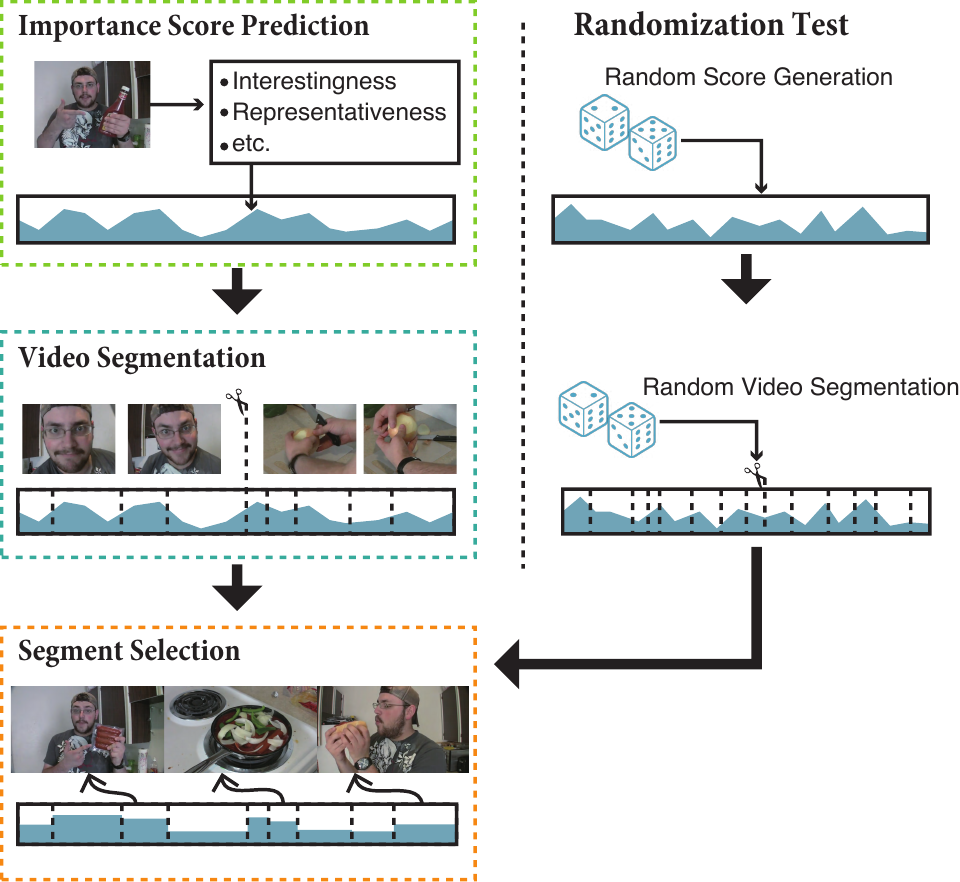}
    \caption{An illustration of the commonly used video summarization pipeline and our randomization test. We utilize random summaries to validate the current evaluation frameworks.}
    \label{fig:sum_pipeline}
\end{figure}

Numerous automatic summarization methods have been proposed in the literature. The most recent methods follow a paradigm that consists of a video segmentation, importance score prediction, and video segment selection as illustrated in Figure \ref{fig:sum_pipeline}. The most challenging part of this pipeline is the importance score prediction, where the task is to highlight the parts that are most important for the video content. Various factors affect importance of video parts, and different video summaries are possible for a single video given a different criterion of importance. In fact, previous works have proposed a variety of importance criteria, such as visual interestingness \cite{Gygli2014,Gygli2015}, compactness (\ie, smaller redundancy) \cite{Zhao2014}, and diversity \cite{Zhang2016,AAAI1816395}. 

\begin{figure}[t!]
\includegraphics[clip, width=\linewidth]{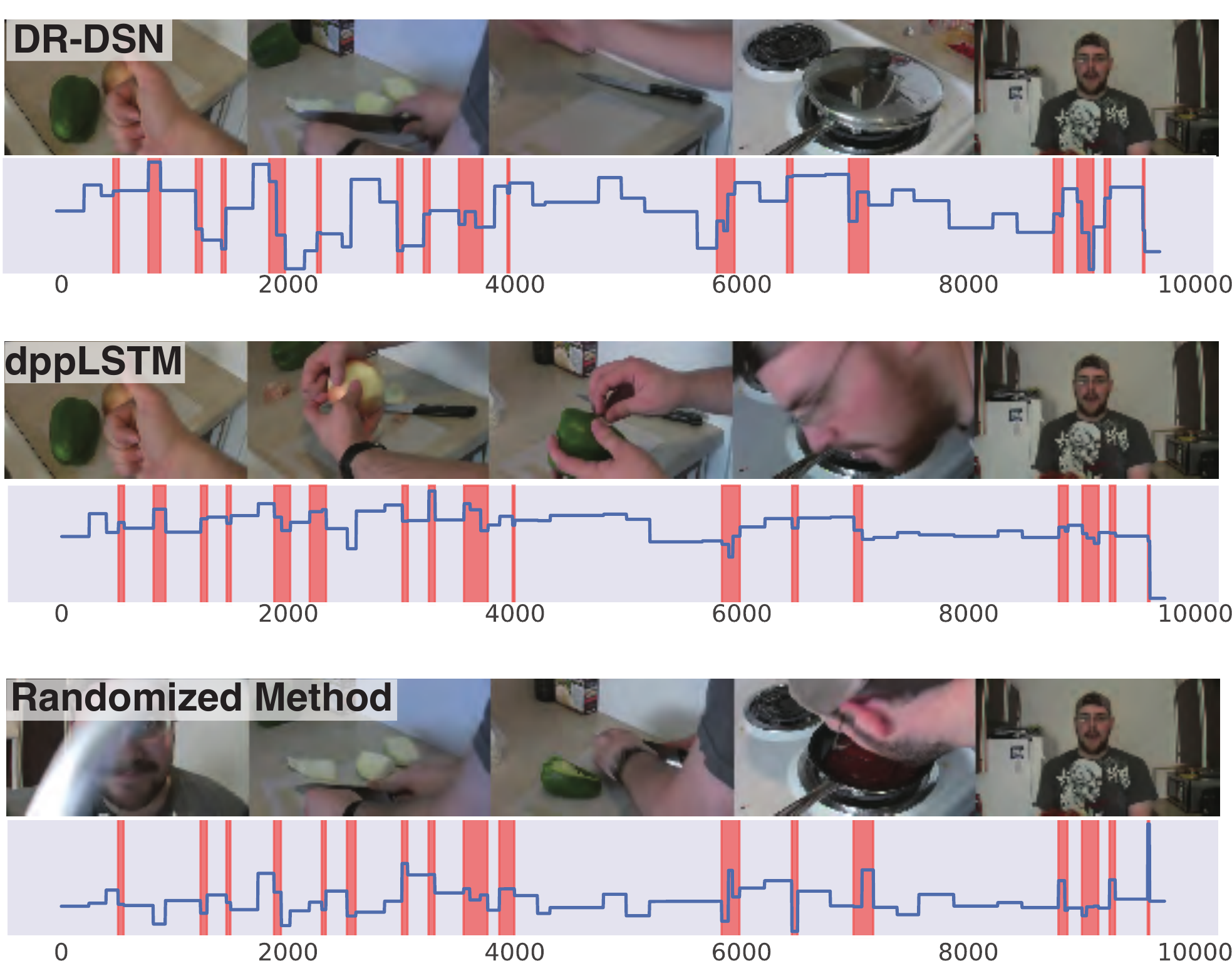}
\caption{Comparison of summaries created by two recent methods and our randomized method (Section~\ref{sec:random_summary}). The blue line shows the segment level importance scores with respect to time (frames). The orange areas indicate the frames selected for the final summary. All of the three methods use the same segment boundaries by KTS \cite{Potapov2014}. Interestingly, all methods (including the random one) produce very similar outputs despite clear differences in the importance scores.}
\label{fig:all_similar}
\end{figure}

Despite the extensive efforts toward automatic video summarization, the evaluation of the generated summaries is still an unsolved problem. A straightforward but yet convincing approach would be to utilise a subjective evaluation; however, collecting human responses is expensive, and reproduction of the result is almost impossible due to the subjectivity. Another approach is to compare generated video summaries to a set of fixed reference summaries prepared by human annotators. To this end, the human annotators are asked to create video summaries, which are then treated as ground truth. The advantage of this approach is the reusability of the reference summaries, \ie, different video summarization methods can be evaluated without additional annotations and the experiments can be reproduced.

The most popular datasets used for reference based evaluations are SumMe \cite{Gygli2014} and TVSum \cite{Song2015}. These datasets provide a set of videos as well as multiple human generated reference summaries (or importance scores) for each original video. The basic evaluation approach, used with both datasets, is to measure the agreement between the generated summary and the reference summaries using F1 score. Since their introduction, SumMe and {TVSum} have been widely adopted in the recent video summarization literature  \cite{Gygli2015,Otani2016b,AAAI1816581,Zhang2016,Zhang_eccv2016,Zhang2018,AAAI1816395}. Nevertheless, the validity of reference summary-based evaluation has not been previously investigated.

This paper delves deeper into the current reference based evaluation framework using SumMe \cite{Gygli2014} and TVSum \cite{Song2015} datasets. We will first review the framework and then apply a randomization test to assess the quality of the results. The proposed randomization test generates video summaries based on random importance scores and random video segmentation. Such summaries provide a baseline score that is achievable by chance.

Figure~\ref{fig:all_similar} illustrates one of the main findings of our work. It turned out that the random method produces summaries that are almost identical to the state-of-the-art despite the fact that it is not using the video content at all for importance score prediction. Deeper analysis shows that while there are differences in the importance scores they are ignored when assembling the final summary. The randomization test revealed critical issues in the current video summarization evaluation protocol, which motivated us to propose a new framework for assessing the importance rankings. 

The main contributions of this paper are as follows:

\begin{itemize}
\item We assess the validity of the current reference summary-based evaluation framework and reveal that a random method is able to reach similar performance scores as the current state-of-the-art. 

\item We demonstrate that the widely used F1 score is mostly determined by the distribution of video segment lengths. Our analysis provides a simple explanation for this phenomenon. 

\item We demonstrate evaluating the importance rankings using correlation between the predicted ordering and the ordering by human annotators. Moreover, we propose several visualisations that give insight to the predicted scoring versus random scores.

\end{itemize}

\begin{table*}[t!]
\centering
\caption{The F1 measures for SumMe and TVSum benchmarks as reported in recent works. Average (Avr) denotes the average of F1 scores over all reference summaries and maximum (Max) denotes the highest F1 score within the reference summaries \cite{Gygli2015}. In addition, we show the F1 values for our randomized test and human annotations (leave-one-out test). It can be noted that random summaries achieve comparable results to the state-of-the-art and even to human annotations.}
\label{tbl:reported_score}
\begin{tabular}{llcccc}
\hline
                 &                         & \multicolumn{2}{c}{SumMe}        & \multicolumn{2}{c}{TVSum} \\
           & Video segmentation      & \multicolumn{1}{c}{Avr.} & Max. & Avr.         & Max.        \\ \hline
LSVS \cite{Khosla2013}            &                         & ---                       & ---  & 0.36         & ---        \\
QRTS \cite{Zhao2014}            & Uniform segmentation    & ---                       & ---  & 0.46         & ---         \\
CSUV \cite{Gygli2014}            & Superframe segmentation & 0.23                      & ---  & ---          & ---         \\
TVSum \cite{Song2015}            & Change-point detection  & ---                       & ---  & 0.50         & ---         \\
VS-LMM \cite{Gygli2015}          & Uniform segmentation    & ---                       & 0.40 & ---          & ---         \\
dppLSTM \cite{Zhang_eccv2016}          & KTS                     & ---                       & 0.43 & 0.60         & ---         \\
VS-DSF \cite{Otani2016b}          & Uniform segmentation    & 0.18                      & ---  & ---          & ---         \\
Summary Transfer \cite{Zhang2016} & KTS                     & ---                       & 0.41 & ---          & ---         \\
DR-DSN \cite{AAAI1816395}        & KTS                     & ---                       & 0.41  & 0.58  & ---        \\ 
re-seq2seq \cite{Zhang2018}      & LSTM-based segmentation & ---                      & 0.45  & 0.64   & ---            \\
SASUM \cite{AAAI1816581}         & KTS                     & ---                        & 0.45  & 0.58  & ---            \\
 \hline 
Randomized test        & KTS                     & 0.19                        & 0.41  & 0.57  & 0.78            \\
Randomized test         & Two-peak   &     0.14    &     0.27     &      0.58    &  0.71    \\
Human        & KTS                     & 0.31                        & 0.54  & 0.54  & 0.78            \\
\hline
\end{tabular}
\end{table*}

\section{Related Work}
\label{sec:related_work}

\subsection{Video Summarization}

A diverse set of video summarization approaches have been presented in the literature. One group of works aim at detecting important shots by measuring the visual interestingness \cite{Gygli2014}, such as dynamics of visual features \cite{Laganiere2008}, and visual saliency \cite{Ma2002}. Gygli \etal~\cite{Gygli2015} combined multiple properties including saliency, aesthetics, and presence of people in the frames.

Another group of methods aims at compactness by discarding redundant shots \cite{Zhao2014}. Maximization of representativeness and diversity in the output video are also widely used criteria in the recent works \cite{Gong2014,Potapov2014,Zhang2016}. These methods are based on the assumption that a good summary should have diverse content while the sampled shots explain the events in the original video.

Recently, LSTM-based deep neural network models have been proposed to directly predict the importance scores given by the human annotators \cite{Zhang_eccv2016}. The model is also extended with determinantal point process \cite{kulesza2012determinantal} to ensure diverse segment selection. Finally, Zhou \etal~\cite{AAAI1816581} applied reinforcement learning to obtain a policy for the frame selection in order to maximize the diversity and representativeness of the generated summary.

Although these works use various importance criteria, many of them employ a similar processing pipeline. Firstly, the importance scores are produced for each frame in the original video. Secondly, the obtained video is divided into short segments. Finally, the output summary is generated by selecting a subset of video segments by maximising the importance scores with the knapsack constraint. 

\subsection{Video Summary Evaluation}

The evaluation of a video summary is a challenging task. This is mainly due to subjective nature of the quality criterion that varies from viewer to viewer and from one time instant to another. The limited number of evaluation videos and annotations further magnify this ambiguity problem.

Most early works \cite{Lu2013,Ma2002,Taskiran2006} as well as some recent works \cite{highlight-detection-with-pairwise-deep-ranking-for-first-person-video-summarization} employ user studies, in which viewers subjectively score the quality of output video summaries prepared solely for the respective works \cite{Lu2013,Sang2010,Yao2016}. The critical shortcoming in such approach is the related cost and reproducibility. That is, one cannot obtain the same evaluation results, even if the same set of viewers would re-evaluate the same videos.

Many recent works instead evaluate their summaries by comparing them to reference summaries. Khosla \etal~\cite{Khosla2013} proposed to use the pixel-level distance between keyframes in reference and generated summaries. Lee \etal~\cite{Lee2012} use number of frames that contain objects of interest as a similarity measure. Gong \etal~\cite{Gong2014} compute precision and recall scores over keyframes selected by human annotators. Yeung \etal~\cite{Yeung} propose a different approach and evaluate the semantic similarity of the summaries based on textual descriptions. For this, they generated a dataset with long egocentric videos for which the segments are annotated with textual descriptions. This framework is mainly used to evaluate video summaries based on user queries \cite{46022,Sharghi2016}. More recently, computing overlap between reference and generated summaries has become the standard framework for video summary evaluation \cite{Gygli2014,Gygli2015,Potapov2014,Song2016,Song2015,Zhao2014}.

This paper investigates the evaluation framework where generated summaries are compared to a set of human annotated references. Currently, there are two public datasets that facilitate this type of evaluation. SumMe \cite{Gygli2014} and TVSum \cite{Song2015} datasets provide manually created reference summaries and are currently the most popular evaluation benchmarks. The SumMe dataset contains personal videos and the corresponding reference summaries collected from 15--18 annotators. The TVSum dataset provides shot-level importance scores for YouTube videos. Most of the literature uses the F1 measure between generated summaries and reference summaries as a performance indicator. Table \ref{tbl:reported_score} shows reported scores for both datasets. The SumMe dataset, which has around 15 different reference summaries, has two possible ways for aggregating the F1 scores: One is to compute an average of F1 measures over all reference summaries, and the other is to use the maximum score. 

\section{Current evaluation framework}

\subsection{SumMe}

SumMe is a video summarization dataset that contains 25 personal videos obtained from the YouTube. The videos are unedited or minimally edited. The dataset provides 15--18 reference summaries for each video. Human annotators individually made the reference summaries so that the length of each summary is less than 15\% of the original video length. For evaluation, generated summaries should be subject to the same constraint on the summary length.

\subsection{TVSum}

TVSum contains 50 YouTube videos, each of which has a title and a category label as metadata. Instead of providing reference summaries, the TVSum dataset provides human annotated importance scores for every two second of each video. For evaluation, the reference summaries, with a predefined length, are generated from these importance scores using the following procedure: Firstly, videos are divided into short video segments, which are the same as in the generated summary. Then, the importance scores within a video segment are averaged to obtain a segment-level importance score. Finally, a reference summary is generated by finding a subset of segments that maximizes the total importance score in the summary. The advantage of this approach is the ability to generate summaries with desired length.

\subsection{Evaluation measure}

The most common evaluation approach is to compute F1 measure between the predicted and the reference summaries. 
Let $y_i\in\{0,1\}$ denote a label indicating which frames from the original video is selected to the summary (\ie~\(y_i = 1\) if the $i$-th frame is selected and otherwise $0$). 
Given similar label \(y^*_i\) for the references summary, the F1 score is defined as
\begin{eqnarray}
\mathrm{F1} = \frac{2 \mathrm{PRE} \cdot \mathrm{REC}}{\mathrm{PRE}+\mathrm{REC}},
\end{eqnarray}
where
\begin{eqnarray}
\mathrm{PRE} = \frac{\sum_{i=1}^{N} y_i \cdot y^*_i}{\sum_{i=1}^{N}y_i} \quad \textnormal{and} \quad 
\mathrm{REC} = \frac{\sum_{i=1}^{N} y_i \cdot y^*_i}{\sum_{i=1}^{N}y^*_i},
\end{eqnarray}
are the frame level precision and recall scores. $N$ denotes the total number of frames in the original video.

In the experiments, the F1 score is computed for each reference summary separately and the scores are summarised either by averaging or selecting the maximum for each video. The former approach implies that the generated summary should include segments with largest number of agreement, while the latter argue that all human annotators provided reasonable importance scores and thus the generated summary should have high score if it matches at least one of the reference summaries.

\section{Randomization test}
\label{sec:random_summary}

Commonly video summarization pipeline consists of three components; importance score estimation, video segmentation, and shot selection (Figure \ref{fig:sum_pipeline}). We devise a randomization test to evaluate the contribution of each part to the final evaluation score. In these experiments we generate video summaries that are independent of video content by utilising random importance scores and random video segment boundaries. Specifically, the importance score for each frame is drawn independently from an uniform distribution $[0, 1]$. When needed, the segment-level scores are produced by average pooling the corresponding frame-level random scores. For video segmentation, we utilise the options defined below.

\textbf{Uniform segmentation} divides the video into segments of constant duration. We used 60 frames in our experiments, which roughly corresponds to 2 seconds (the frame rates in SumMe and TVSum datasets are 30 fps and 25 fps, respectively).

\textbf{One-peak segmentation} samples the number of frames in each segment from an unimodal distribution. We assume that the number of frames between adjacent shot boundaries follow the Poisson distribution with event rate $\lambda = 60$.

\textbf{Two-peak segmentation} is similar to one-peak version, but utilises bimodal distribution, \ie, a mixture of two Poisson distributions, whose event rates are $\lambda=30$ and $\lambda=90$, respectively. For sampling, we randomly choose one of the two Poisson distributions with the equal probability and then sample the number of frames. Consequently, a video is segmented into both longer and shorter segments, yet the expected number of frames in a segment is 60 frames.

In addition to the completely random methods, we assess one commonly used segmentation approach and its variation in conjunction with the random scores. 

\textbf{Kernel temporal segmentation (KTS)} \cite{Potapov2014} is based on the visual content of a video and is the most widely used method in the recent video summarization literature (Table~\ref{tbl:reported_score}).
KTS produces segment boundaries by detecting changes in visual features. A video segment tends to be long if visual features do not change considerably.

\textbf{Randomized KTS} first segments the video with KTS and then shuffles the segment ordering; therefore, the distribution of segment lengths is exactly the same as KTS's, but the segment boundaries are not synchronized with the visual features.

F1 scores obtained by these randomized (and partially randomized) summaries serve as a baseline that can be achieved completely by chance. Reasonable evaluation framework should produce higher scores for methods that are producing sensible importance scores. Furthermore, one would expect that human generated ground truth summaries should produce top scores in leave one out experiments.

\begin{figure*}[t!]
\begin{tabular}{c}
\begin{minipage}[b]{.65\linewidth}
\centering
\includegraphics[clip, width=1\linewidth]{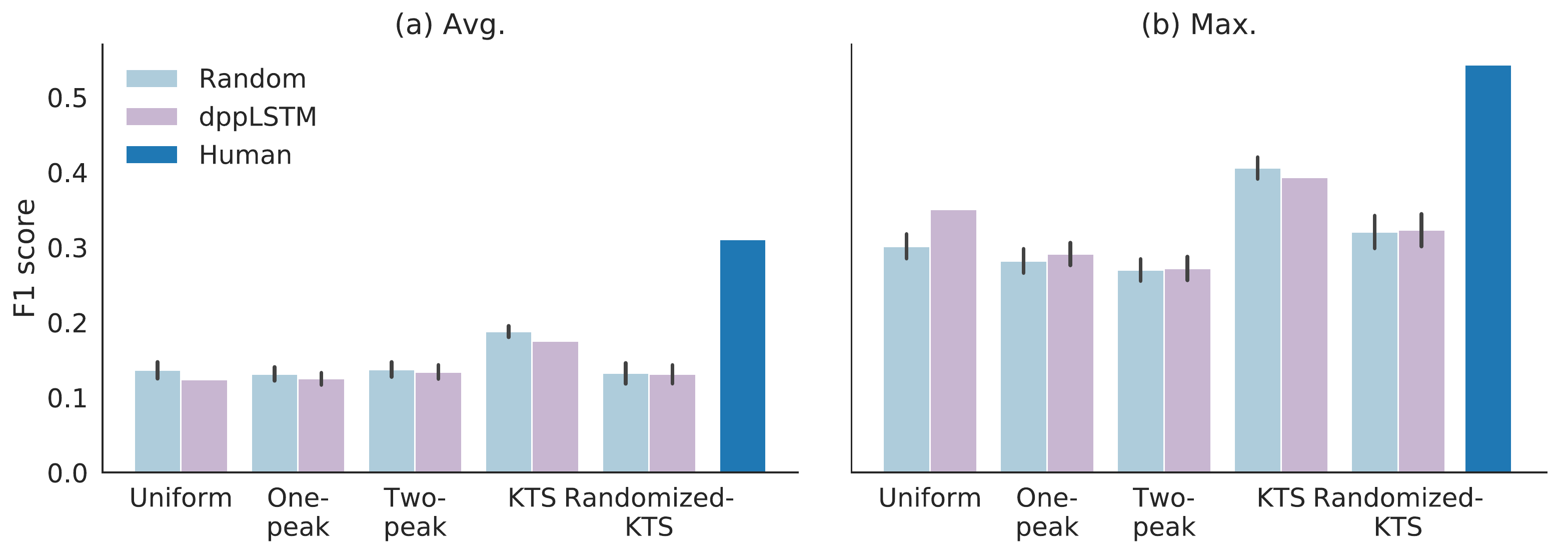}
\caption{F1 scores for different segmentation and importance score combinations for SumMe. Light blue bars refer to random summaries and dark blue bars indicate scores of manually created reference summaries (leave-one-out test). Purple bars show the scores for DR-DSN importance scoring with different segmentation methods. Left: the average of mean F1 scores over reference summaries. Right: the average of the maximum scores.}
\label{fig:summe_random_eval}
\end{minipage}

\hspace{2mm}

\begin{minipage}[b]{.3\linewidth}
\centering
\includegraphics[clip, width=1\linewidth]{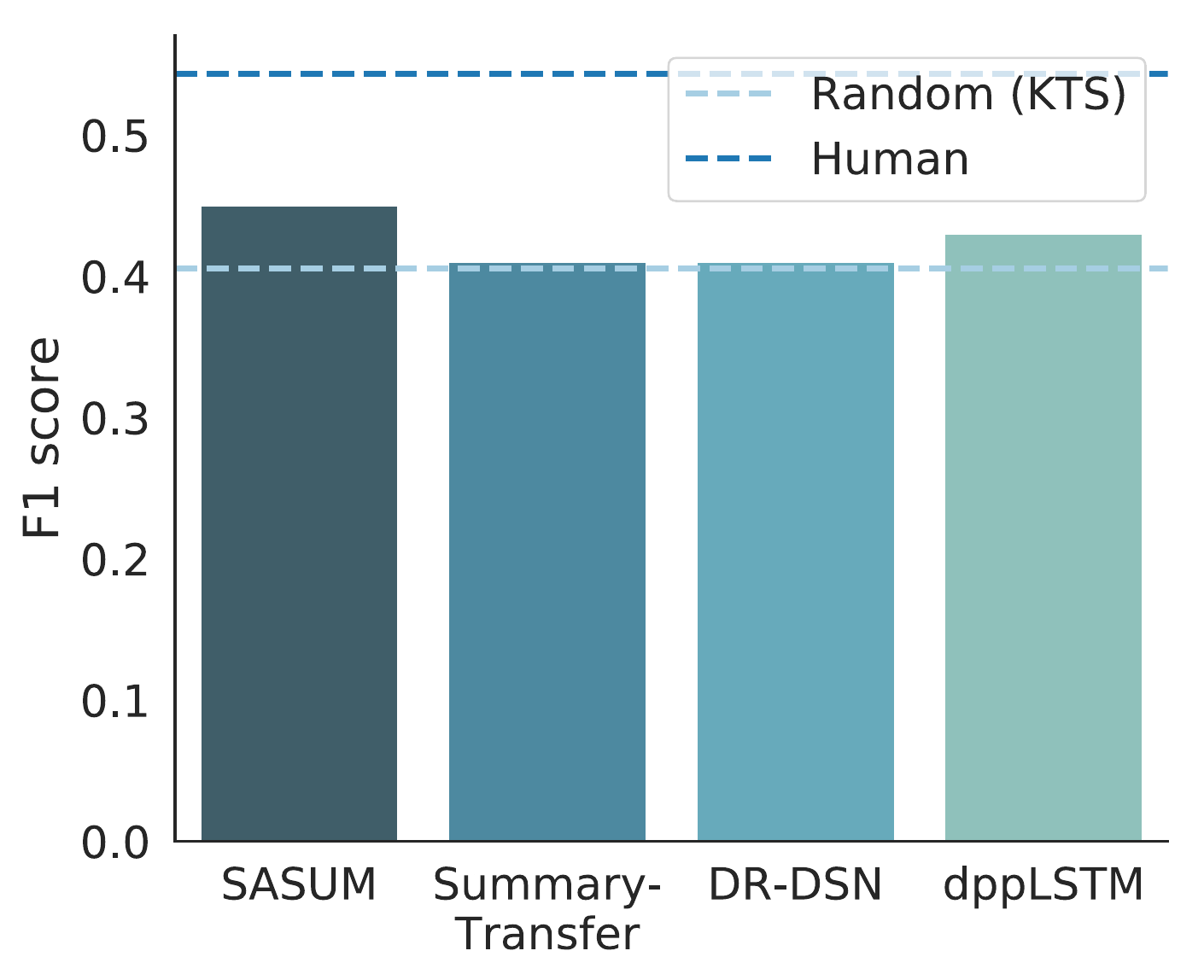}
\caption{Recently reported F1 scores for methods using KTS segmentation in SumMe. The average score for random summaries with KTS segmentation is represented by a light blue dashed line.}
\label{fig:summe_previous_methods}
\end{minipage}
\end{tabular}
\end{figure*}

\subsection{Analysis on the SumMe dataset}
\label{sec:summe}

Figure \ref{fig:summe_random_eval} displays the F1 scores (average and maximum) obtained with different versions of our randomized method (see previous section).
We performed 100 trials for every random setting and the black bar is the 95\% confidence interval.
In addition, the same figure contains the corresponding F1 scores for each random segmentation method, but using frame level importance scores from one recently published methods DR-DSN \cite{AAAI1816395}. The reference performance is obtained using human created reference summaries in leave-one-out scheme. In this case, the final result is calculated by averaging the F1 scores (avg or max) obtained for each reference summary.

Interestingly, we observe that the performance is clearly dictated by the segmentation method and there is small (if any) impact on the importance scoring. Moreover, the difference between human performance and the best performing automatic method is similar in magnitude to the differences between the segmentation approaches. Figure \ref{fig:summe_previous_methods} illustrates the recent state-of-the-art results for SumMe dataset. Surprisingly, KTS segmentation with random importance scores obtains comparable performance to the best published methods. Section~\ref{sec:explanation} provides possible explanations for this phenomenon. 

\subsubsection{Human Evaluation on SumMe}
We conducted human evaluation to compare summaries on the SumMe dataset.
Subjects compare two video summaries and determine which video better summarizes the original video.
In the first experiment, we asked subjects to rate video summaries generated using random importance scores and DR-DSN scores. Both methods use KTS segmentation.
Overall, random scores got a slightly higher score than DR-DSN, however, 46\% of answers were that the summaries are equally good (bad).
This result agree with the observation in the Section~\ref{sec:summe} that the importance scoring hardly affects the evaluation score on the SumMe dataset.
We also compare KTS and uniform segmentation with random importance scoring.
As a result, subjects prefer uniform segmentation for videos recording long activity, \eg, sightseeing of the statue of liberty and scuba diving.
On the other hand, KTS works better for videos with notable events or activities.
For such videos, the important parts have little ambiguity, therefore the F1 scores based on the agreement between generated summaries and reference summaries can get higher.
For the detailed results of the human evaluation, see the supplementary material.

\subsection{Analysis on TVSum dataset}
\label{sec:analysis_tvsum}

\begin{figure*}[t!]
\centering
\includegraphics[clip, width=\linewidth]{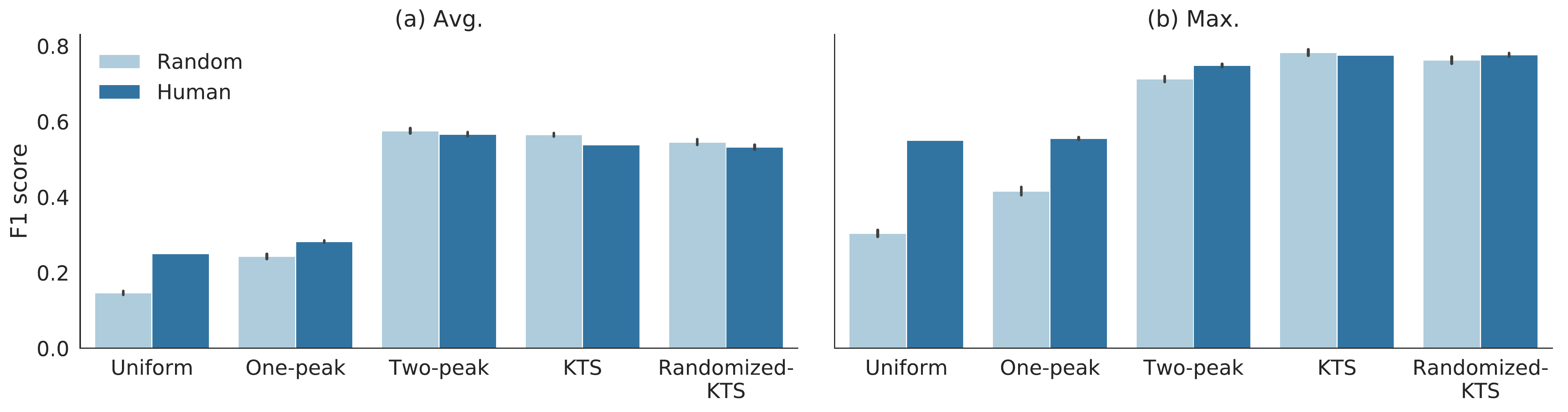}
\caption{F1 scores for different segmentation methods combined to either random or human annotated importance scores (leave-one-out) for TVSum dataset. Light blue bars refer to random scores and dark blue bars indicates human annotations. Interestingly, the random and human annotations obtain similar F1 scores in most cases.}
\label{fig:tvsum_const_eval}
\end{figure*}

Instead of reference summaries, TVSum dataset contains human annotated importance scores for every 2 second segment in the original video. The main advantage of this approach is the ability to generate reference summaries of arbitrary length. It is also possible to use different segmentation methods. For these reasons, TVSum provides an excellent tool for studying the role of importance scoring and segmentation in the current evaluation framework. 

Figure \ref{fig:tvsum_const_eval} displays the F1 scores for different segmentation methods using both random and the human annotated importance scores. In the latter case, the results are computed using leave-one-out procedure. Surprisingly, for the most of the segmentation methods, the random importance scores have similar performance as human annotations. In addition, the completely random two-peak segmentation performs equally well as content based KTS segmentation. Furthermore, the results in Table \ref{tbl:reported_score} illustrate that our random results are on-par (or at least comparable) with the best reported results in the literature. The uniform and one-peak segmentation do not reach the same results, but in these cases the better importance scoring seems to help. In general, the obtained results highlight the challenges in utilizing the current F1 based evaluation frameworks.

\subsection{Discussion}
\label{sec:explanation}

As observed in the previous sections, the random summaries resulted in surprisingly high performance scores. The results were on-par with the state-of-the-art and sometimes surpassed even the human level scores. In particular, the segmentation methods that produce large variation in the segment length (\ie two-peak, KTS, and randomized KTS) produced high F1 scores. The results may be understood by examining how the segment length affects on selection procedure in the knapsack formulation that is most commonly adopted in video summarization methods.

\begin{figure}
    \centering
    \includegraphics[clip,width=\linewidth]{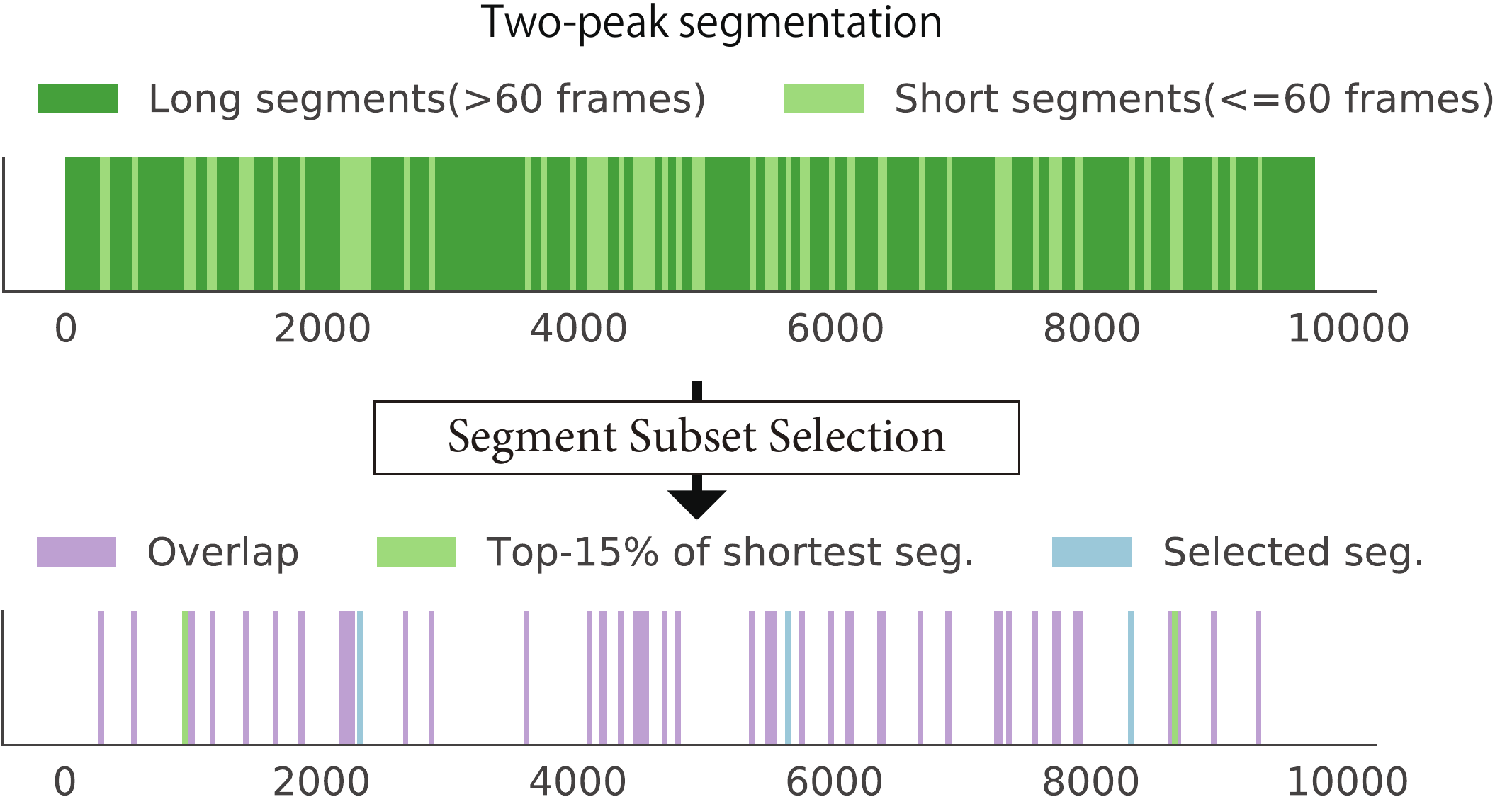}
    \caption{Long segments are implicitly discarded from the summary and only short segments are selected. Top: Green and light green areas visualize segment boundaries generated by the two-peak segmentation method. Bottom plot shows segments selected by dynamic programming algorithm (blue), and top 15\% of the shortest video segments (light green), and segments overlapping between them (Purple). Notice that the most of the selected parts are within the group of the shortest segments.}
    \label{fig:illustrative}
\end{figure}

A dynamic programming solver, commonly used for the knapsack problem, selects a segment only if the corresponding effect on the overall score is larger than that of any combination of remaining segments whose total length is shorter. In other words, a segment A is selected only if there are no segments B and C whose combined length is less than A and the effect to total score is more or equal to A. This is rarely true for longer segments in the current summarization tasks, and therefore the summary is usually only composed of short segments. This phenomenon significantly limits the reasonable choices available for segment subset selection. For example, two-peak segmentation draws a segment length from two distributions whose modes are 30 frames and 90 frames; therefore, we can roughly say that longer segments occupies two-third of the total length. If these longer segments are all discarded, the generated summary only consists of the rest one-third of the original video.For generating a summary whose length is 15\% of the original video duration, most of the segments are expected to be shared for generated and reference summaries regardless of associated importance scores. This is illustrated in Figure \ref{fig:illustrative}. Due to the same reason, the importance scores have more impact if all the segments have equal length (see uniform and one-peak results in Figure \ref{fig:tvsum_const_eval}).

Using the sum of frame-level scores may alleviate the challenge; however, most works instead employ averaging because this drastically increases F1 scores on TVSum. With summation, human summary clearly outperforms random ones, but we can still see the effect of segmentation.

The results on SumMe dataset in Section \ref{sec:summe} illustrate another challenge. For this dataset, KTS-based references obtain really high performance scores. The use of KTS implicitly incorporate small-redundancy strategy, which aims to create a visually non-redundant video summary. That is, KTS groups visually similar frames into a single segment. Therefore, long segments are likely to be redundant and less lively and thus they are less interesting. Human annotators would not like to include such segments in their summaries. Meanwhile, the dynamic programming-based segment subset selection tends to avoid long segments as discussed above. Thus generated summaries tend to match the human preference.

\section{Importance score evaluation framework}
The aforementioned challenges render the current benchmarks inapplicable for assessing the quality of the importance scores. At the same time, most of the recent video summarization literature present methods particularly for importance score prediction. To overcome this problem, we present a new alternative approach for the evaluation.

\subsection{Evaluation using rank order statistics}
\begin{figure}[t!]
\centering
\includegraphics[clip, width=\linewidth]{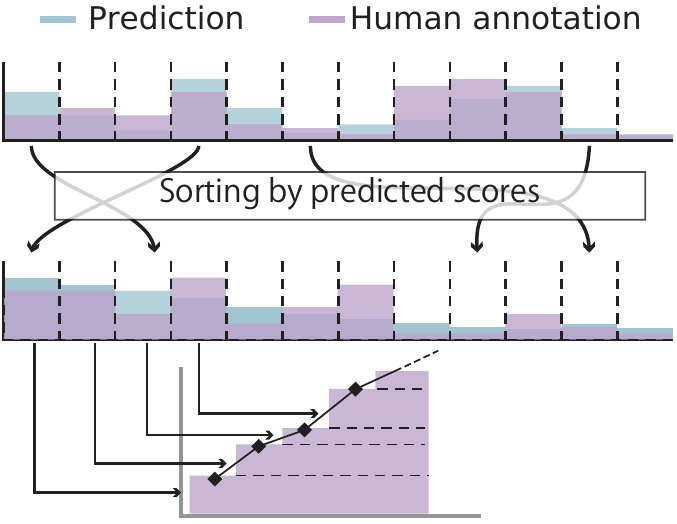}
\caption{Overview of the score curve formation.
}
\label{fig:accum_curve}
\end{figure}
In statistics, rank correlation coefficients are well established tools for comparing the ordinal association (i.e.~relationship between rankings). We take advantage of these tools in measuring the similarities between the implicit rankings provided by generated and human annotated frame level importance scores as in \cite{Vasudevan2017}.

More precisely, we use Kendall's \(\tau\) \cite{kendall1945treatment} and Spearman's $\rho$ \cite{kokoska1999crc} correlation coefficients. To obtain the results, we first rank the video frames according to the generated importance scores and the human annotated reference scores (one ranking for each annotator). In the second stage, we compare the generated ranking with respect to each reference ranking. The final correlation score is then obtained by averaging over the individual results. 

We demonstrate the rank order correlation measures, by evaluating two recent video summarization methods (dppLSTM \cite{Zhang_eccv2016} and DR-DSN \cite{AAAI1816395}). For both methods, we utilise the implementations provided by the original authors. For sanity check, we also compute the results using random scoring, which by definition should produce zero average score. These results are obtained by generating 100 uniformly-distributed random value sequences in $[0, 1]$ for each original video and averaging over the obtained correlation coefficients. The human performance is produced using leave-one-out approach. Table~\ref{tbl:rank_order_statistics} summarizes the obtained results for TVSum dataset. 

Overall, the metric shows a clear difference between tested methods and the random scoring. In addition, the correlation coefficient for human-annotations is significantly higher than for any other method, which confirms that human importance scores correlate to each other. From the tested methods, dppLSTM results in higher performance compared to DR-DSN. This makes sense, since dppLSTM is particularly trained to predict human annotated importance scores, while DR-DSN aims at maximizing the diversity of the content in the generated summaries. However, both methods clearly outperform the random scoring. 

We further investigate the relation between the correlation measures for importance scores and the quality of output video summaries.
We compare video summaries generated using importance scores which positively correlate with human annotations and those using importance scores with negative correlation.
The result of human evaluation demonstrated that video summaries generated using importance scores with positive correlation outperformed others.
The details of the result are in the supplementary material.

\begin{table}[t!]
\centering
\caption{Kendall's \(\tau\) and Spearman's $\rho$ correlation coefficients computed between different importance scores and manually annotated scores on TVSum dataset.}
\label{tbl:rank_order_statistics}
\begin{tabular}{lcc}
\hline
Method & Kendall's $\tau$ & Spearman's $\rho$ \\ \hline
dppLSTM \cite{Zhang_eccv2016}& 0.042             & 0.055             \\
DR-DSN \cite{AAAI1816395}& 0.020             & 0.026             \\
Random                  & 0.000             & 0.000             \\
Human                   & 0.177             & 0.204             \\ \hline
\end{tabular}
\end{table}

\subsection{Visualizing importance score correlations}

One of the main challenges in the evaluation of video summaries is the inconsistency between the human-annotations. In fact, although the human annotations result in the highest correlation coefficient in Table \ref{tbl:rank_order_statistics}, the absolute value of the correlation is still relatively low. This stems from subjectivity and ambiguity in the importance score annotation. As we can imagine, what is important in a video can be highly subjective, and the annotators may or may not agree. Furthermore, even if the annotators agree that a certain video content is important, there can be multiple parts in a video that contain the same content in different viewpoints and expressions. Selection from these parts may still be ambiguous problem.

To highlight the variation in the annotations, we propose to visualize the predicted importance score ranking with respect to the reference annotations. To do this, we first compute the frame level average scores over the human annotators. In the second stage, we sort the frames with respect to the predicted importance scores in descending order (Figure \ref{fig:accum_curve}, middle). In the final step, we accumulate the averaged reference scores based on the ranking obtained in the second stage. More precisely,  
\begin{equation}
a_i = \sum_{t=1}^i \frac{s_t}{\sum_{j=1}^n s_j},
\end{equation}
where $s_i$ denotes the average human-annotated score for the $i$-th frame in the sorted video. The normalization factor in the denominator ensures that the maximum value equals to $1$. As shown in Figure \ref{fig:accum_curve} (bottom), $a_i$ forms a monotonically increasing curve over the sorted frames. If the predicted scores have high correlation to human scores, the curve should increase rapidly. Similar curves can be produced for the human scores using leave-one-out approach.  

\begin{figure}
\centering
\begin{tabular}{c}

\begin{minipage}{0.5\hsize}
\centering
\includegraphics[clip, width=\linewidth]{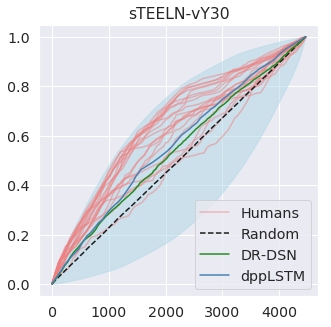}
(a)
\end{minipage}

\begin{minipage}{0.5\hsize}
\centering
\includegraphics[clip, width=\linewidth]{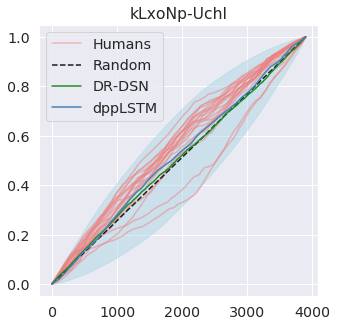}
(b)
\end{minipage}

\end{tabular}
\caption{Example correlation curves produced for two videos from TVSum dataset (sTEELN-vY30 and kLxoNp-UchI are video ids). The red lines represent correlation curves for each human annotator and the black dashed line is the expectation for a random importance scores. The blue and green curves show the corresponding results to dppLSTM and DR-DSN methods, respectively. See supplementary material for more results.  }
\label{fig:acc_score_curve}
\end{figure}

Figure~\ref{fig:acc_score_curve} shows correlation curves produced for two videos from TVSum dataset. The red lines show the $a_t$ curve for each human annotator and the black dashed line is the expectation for a random importance scores. The blue and green curves show the corresponding results to dppLSTM and DR-DSN methods, respectively. The light-blue colour illustrates the area, where correlation curves may lie. That is, when the predicted importance scores are perfectly concordant with averaged human-annotated scores, \ie, the score based rankings are the same, the curve lies on the upper bound of the light-blue area. On the other hand, a curve coincides with the lower bound of the the area when the ranking of the scores is in a reverse order of the reference. 

\begin{figure}[t!]
\centering
\includegraphics[clip, width=\linewidth]{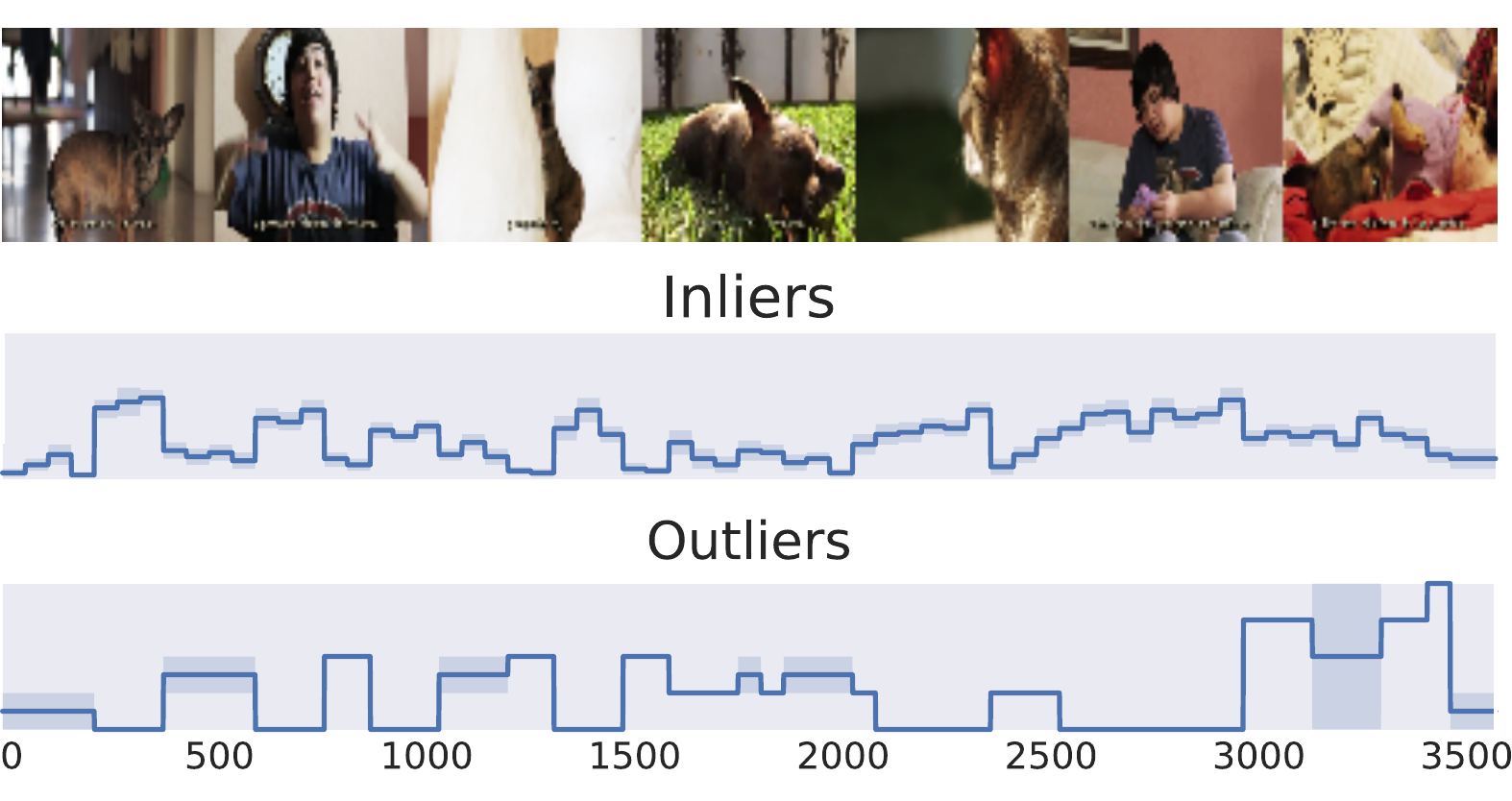}
\caption{Comparison of human-annotated scores. The bottom row shows the frame level importance scores for the selected two human annotators (outliers). The middle row displays the similar score obtained by averaging over the remaining human annotators (inliers). The top row illustrates keyframes from the corresponding video. One can notice that inliers and outlier have highlighted almost completely opposite parts of the video.}
\label{fig:outlier_scores}
\end{figure}

The most of the human annotators obtain a curve that is well above the random baseline in Figure~\ref{fig:acc_score_curve}. Moreover, Figure~\ref{fig:acc_score_curve} (a) shows that both dppLSTM and DR-DSN are able to predict importance scores that are positively correlated with human annotations. On the other hand, Figure~\ref{fig:acc_score_curve} (b) shows two red lines that are well below the black dashed line. This implies that these annotators labelled almost opposite responses to the overall consensus. Detailed observation in Figure \ref{fig:outlier_scores} reveals that this is indeed the case. The outliers highlighted segments around 1500 and 3000 frames, on the other hand, other annotators showed almost opposite opinion for the segments. The proposed visualization provides intuitive tool for illustrating such tendencies.

\section{Conclusion}

Public benchmark datasets play an important role as they facilitating easy and fair comparison of methods. The quality of the benchmark evaluations have high impact as the research work is often steered to maximise the benchmark results. In this paper, we have assessed the validity of two widely used video summarization benchmarks. Our analysis reveals that the current F1 score based evaluation framework has severe problems. 

In most cases it turned out that randomly generated summaries were able to reach similar or even better performance scores than the state-of-the-art methods. Sometimes the performance of completely random method surpassed that of human annotators. Closer analysis revealed that score formation was mainly dictated by the video segmentation and particularly the distribution of the segment lengths. This was mainly due to the widely used subset selection procedure. In most cases, the contribution of the importance scores were completely ignored by the benchmark tests. 

Based on our observations, we proposed to evaluate the methods using the correlation between predicted and human-annotated importance scores instead of the final summary given by the segment subset selection process. The introduced evaluation offers additional insights about the behaviour of the summarization methods. We also proposed to visualize the correlations by accumulative score curve, which intuitively illustrates the quality of the importance scores with respect to various human annotations. 

The proposed new evaluation framework covers only methods that estimate the frame level importance scores. It is not suitable for other approaches such as \eg, clustering-based methods that pick out video segments close to cluster centres. In addition, we primarily addressed the evaluation based on correlation with human annotations. Other factors like comprehensibility of a story in a video, visual aesthetics and relevance to a user query would also be valuable for various applications. We believe that it would be important to address these aspects in future works. Moreover, we believe that new substantially larger datasets are needed for pushing video summarization research forward.

\textbf{Acknowledgement}
This work was partly supported by JSPS KAKENHI Grant Nos.~16K16086 and 18H03264.

{\small
\bibliographystyle{ieee}
\bibliography{egbib}
}

\newpage

\section*{Appendix}
\subsection*{Distribution of Segment Length}
\begin{figure}[h!]
    \centering
    \includegraphics[clip, width=\linewidth]{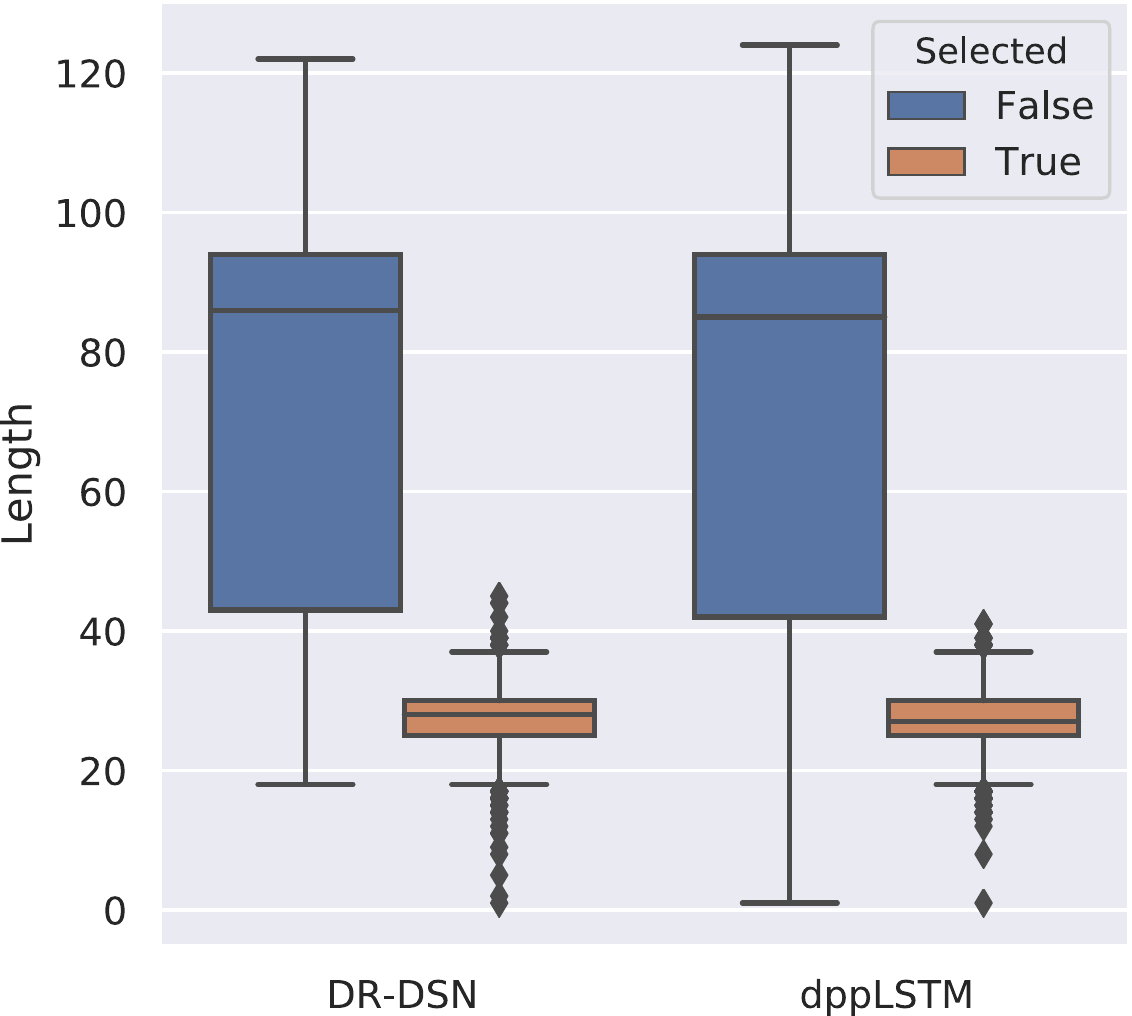}
    \caption{Box plot of segment length. Orange boxes represent segments selected for output summaries, and blue boxes indicate remaining ones.}
    \label{fig:seg_length}
\end{figure}

Figure~\ref{fig:seg_length} shows box plots of the segment lengths. The orange boxes refer to the segments selected for video summaries and blue boxes indicate the remaining ones.
The segment boundaries are generated by two-peak segmentation in Section 4.
As described in Section 4.3, Figure~\ref{fig:seg_length} shows that segment selection results in summaries composed of only short segments.
The distribution of segment lengths is similar for DR-DSN and dppLSTM.
This indicates that importance score prediction methods hardly affect segment selection.

\subsection*{Examples of Video Summaries}
\begin{figure*}
\centering
\vspace{10mm}
\begin{tabular}{cc}
\begin{minipage}{0.5\hsize}
\centering
dppLSTM\\
\includegraphics[clip,width=\linewidth]{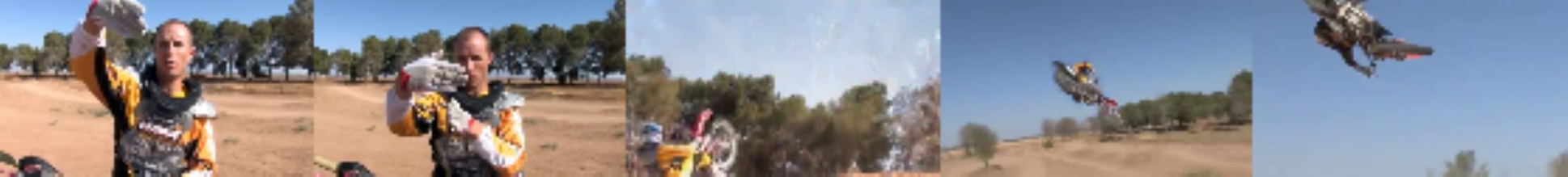} 
\includegraphics[clip,width=\linewidth]{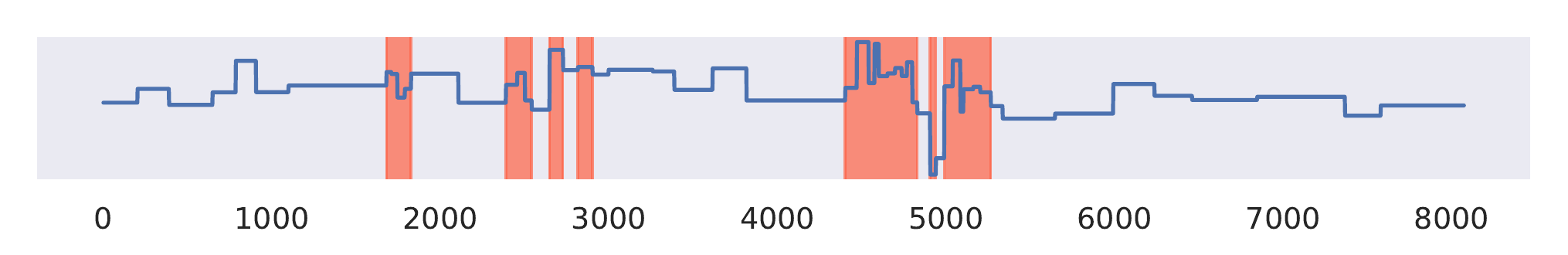}
\\
DR-DSN\\
\includegraphics[clip,width=\linewidth]{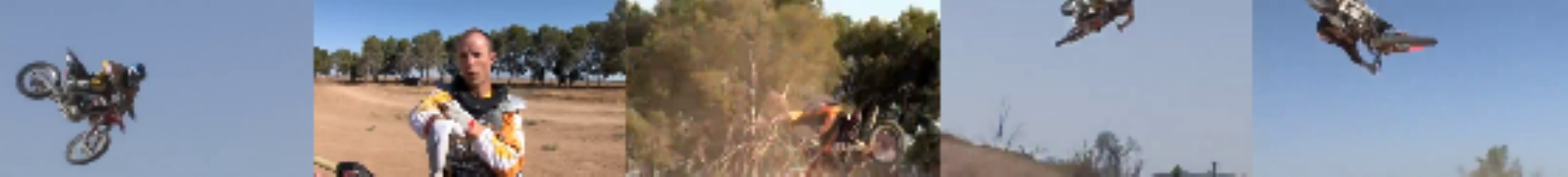} 
\includegraphics[clip,width=\linewidth]{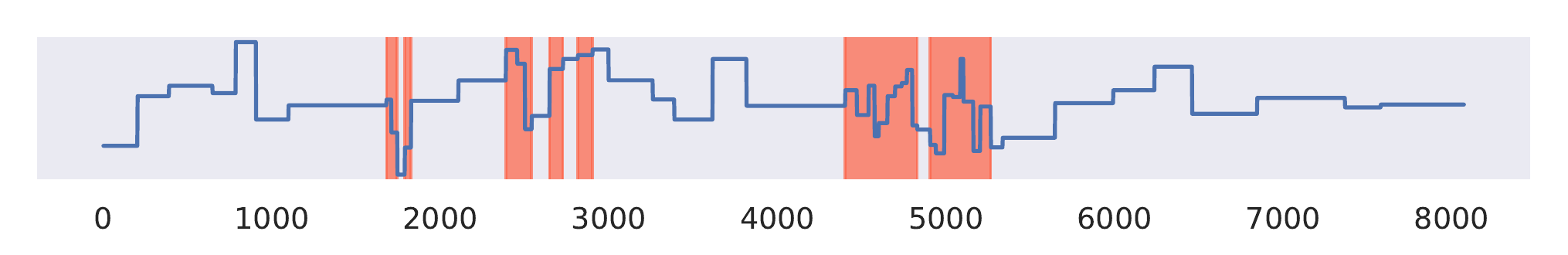}
\\
Randomized method\\
\includegraphics[clip,width=\linewidth]{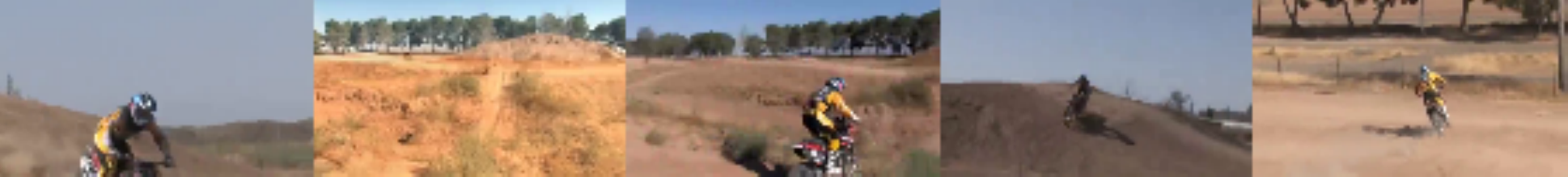}
\includegraphics[clip,width=\linewidth]{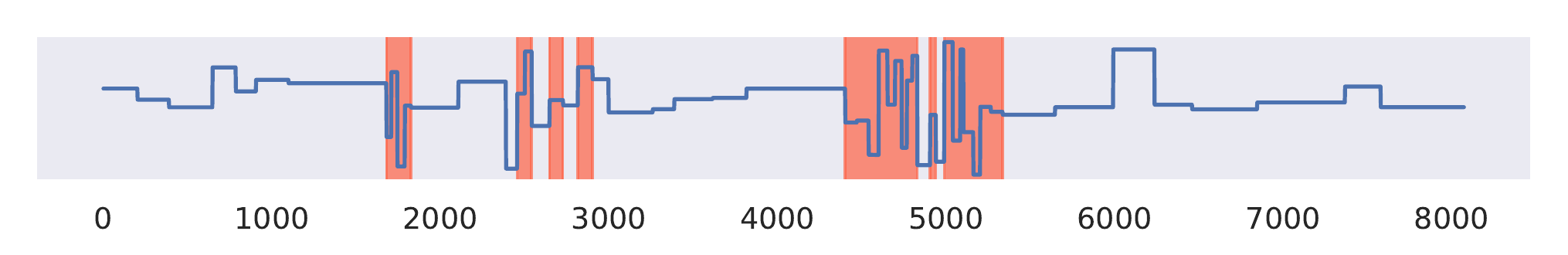}
\end{minipage}

&

\begin{minipage}{0.5\hsize}
\centering
dppLSTM\\
\includegraphics[clip,width=\linewidth]{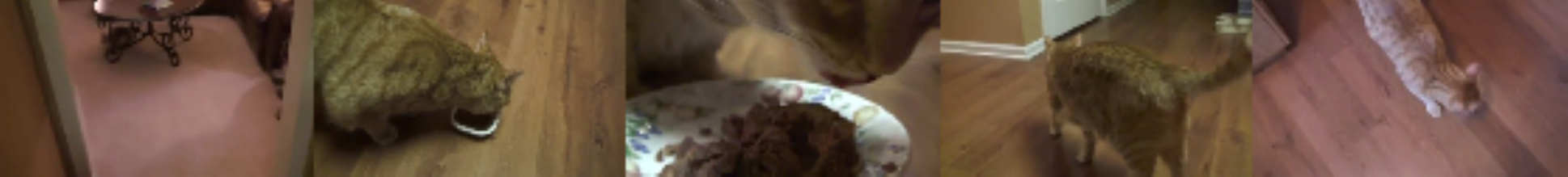} 
\includegraphics[clip,width=\linewidth]{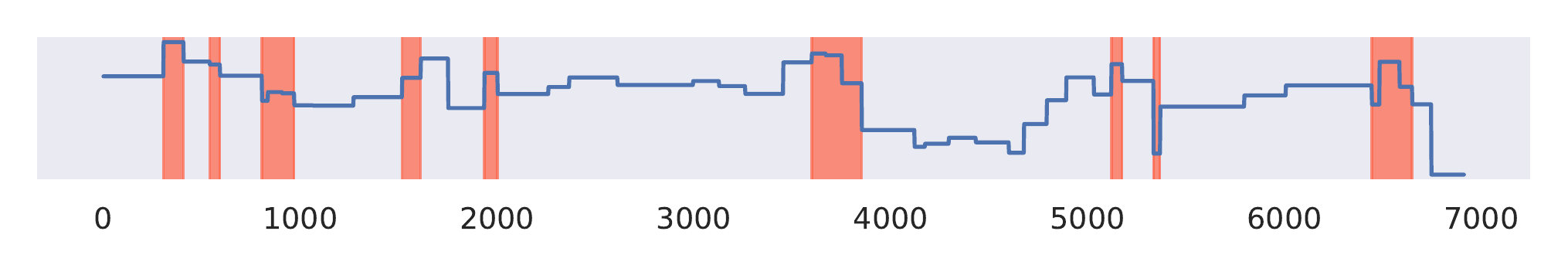}
\\
DR-DSN\\
\includegraphics[clip,width=\linewidth]{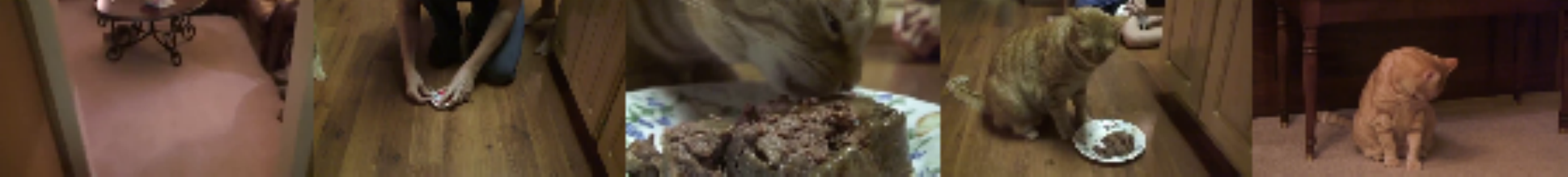}
\includegraphics[clip,width=\linewidth]{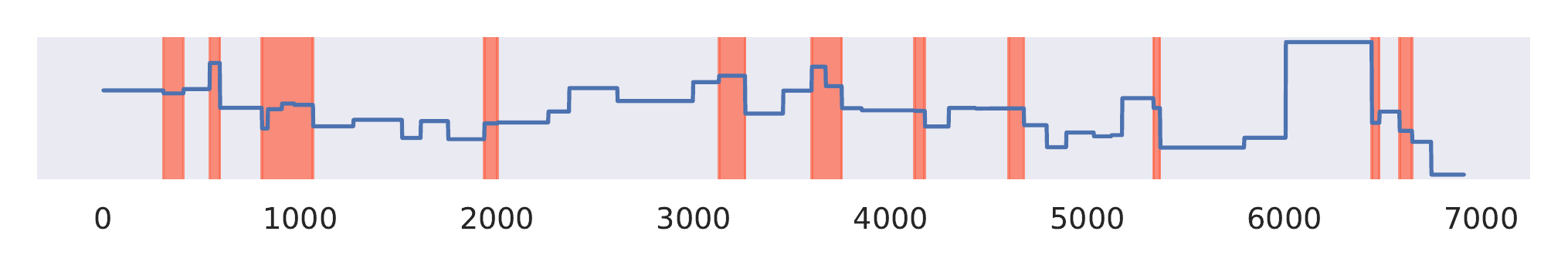}
\\
Randomized method\\
\includegraphics[clip,width=\linewidth]{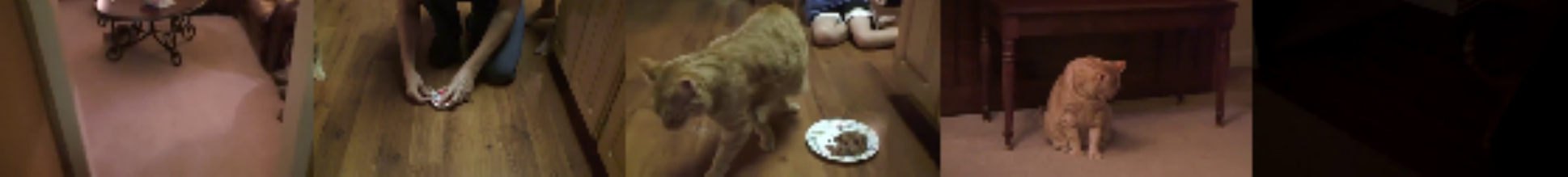}
\includegraphics[clip,width=\linewidth]{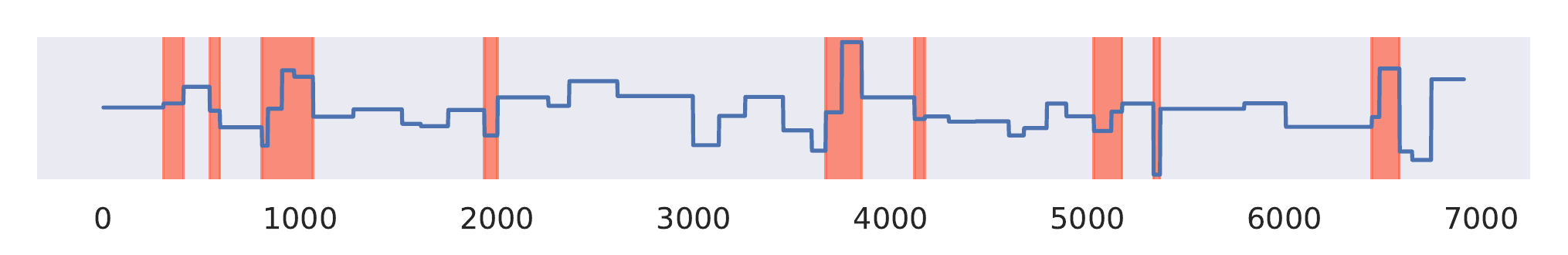}
\end{minipage}

\\
\\
\\
\\

\begin{minipage}{0.5\hsize}
\centering
dppLSTM\\
\includegraphics[clip,width=\linewidth]{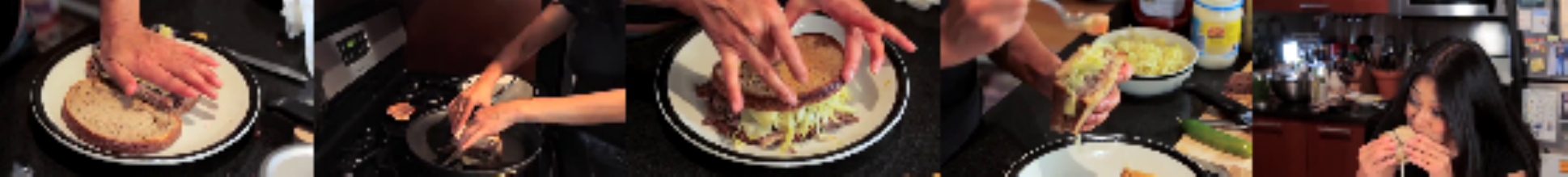} 
\includegraphics[clip,width=\linewidth]{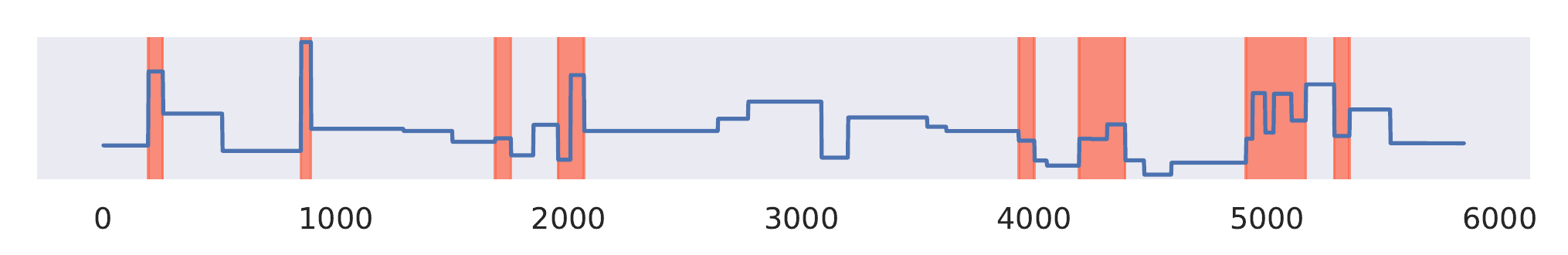}
\\
DR-DSN\\
\includegraphics[clip,width=\linewidth]{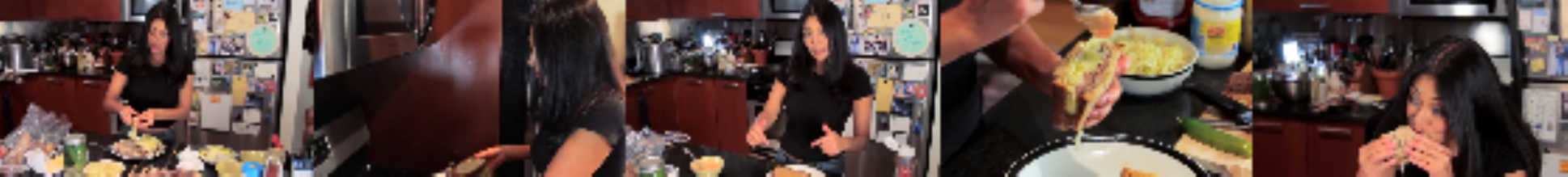}
\includegraphics[clip,width=\linewidth]{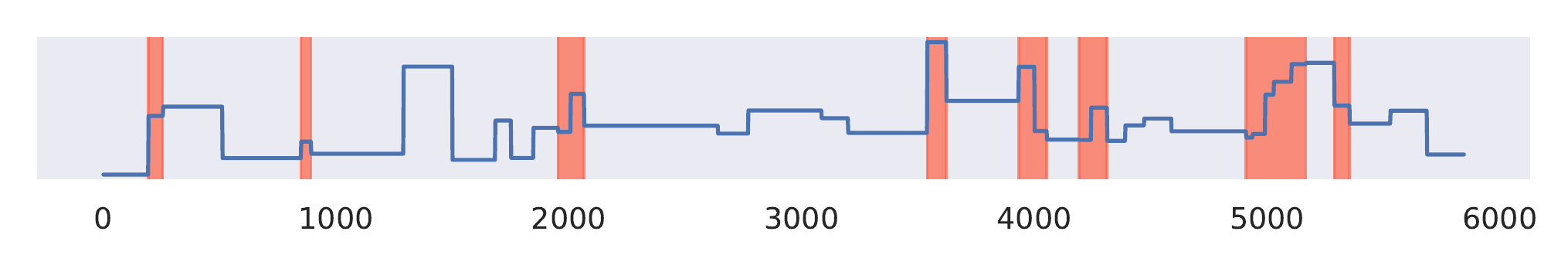}
\\
Randomized method\\
\includegraphics[clip,width=\linewidth]{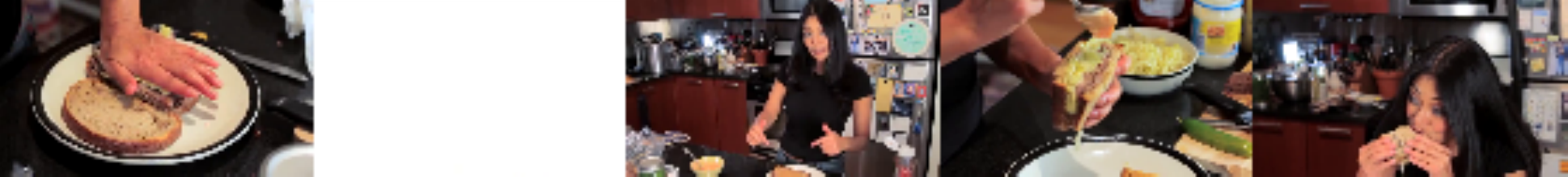}
\includegraphics[clip,width=\linewidth]{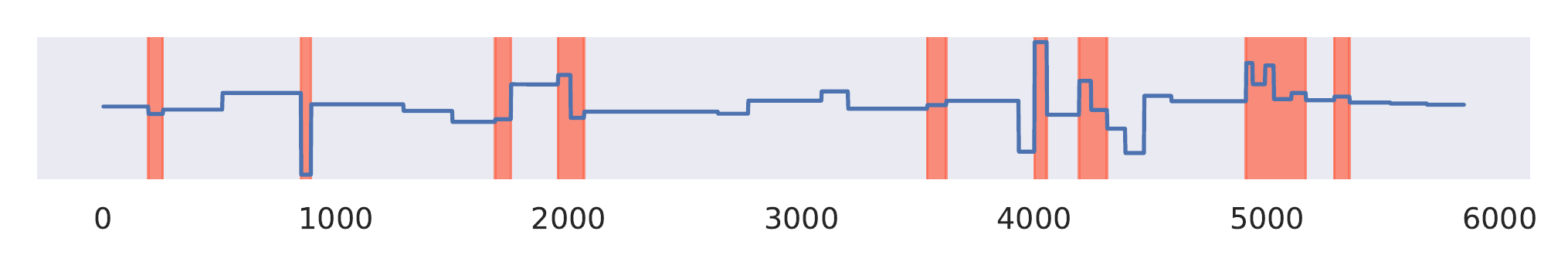}
\end{minipage}

&

\begin{minipage}{0.5\hsize}
\centering
dppLSTM\\
\includegraphics[clip,width=\linewidth]{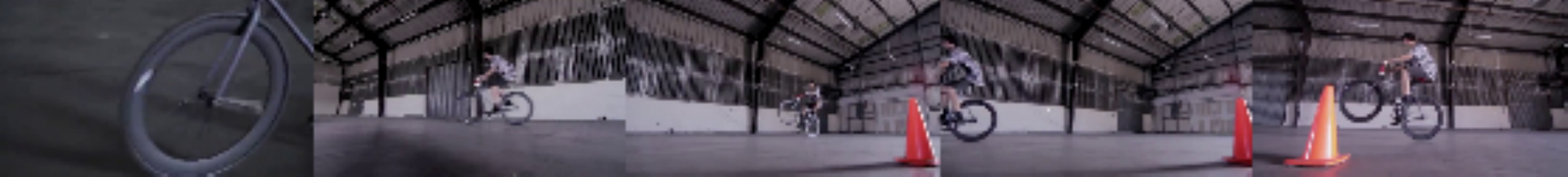} 
\includegraphics[clip,width=\linewidth]{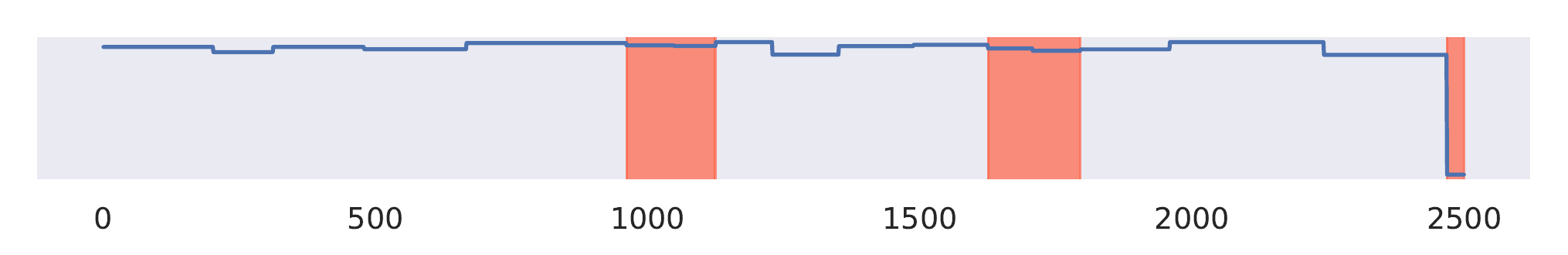}
\\
DR-DSN\\
\includegraphics[clip,width=\linewidth]{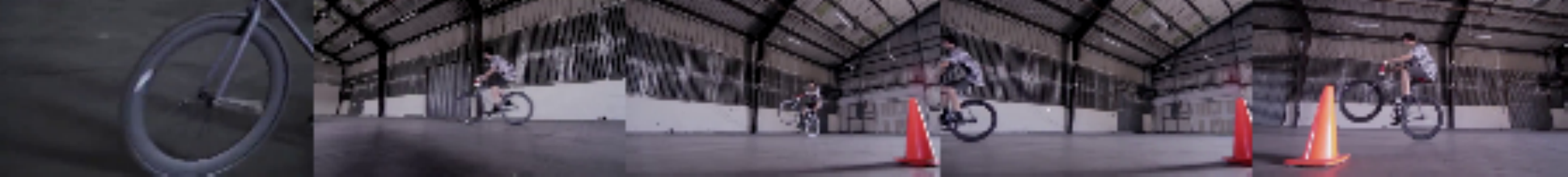}
\includegraphics[clip,width=\linewidth]{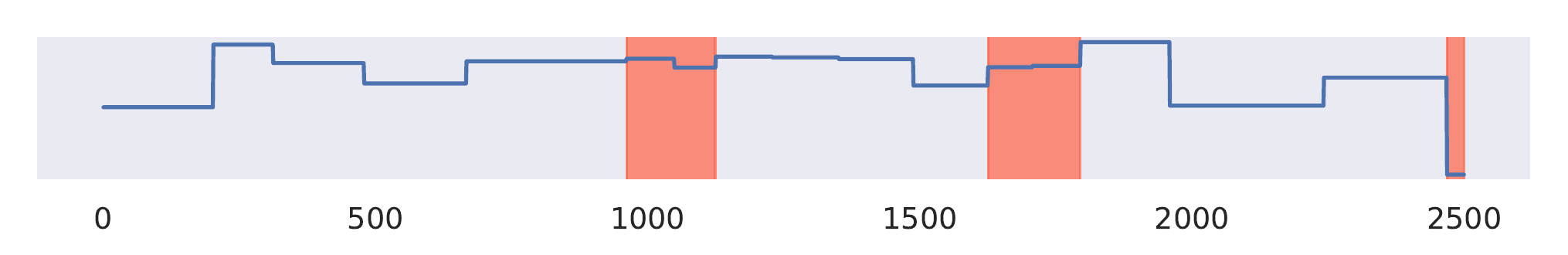}
\\
Randomized method\\
\includegraphics[clip,width=\linewidth]{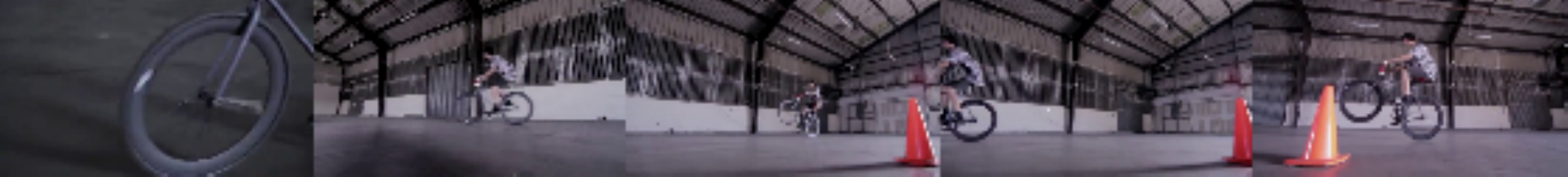}
\includegraphics[clip,width=\linewidth]{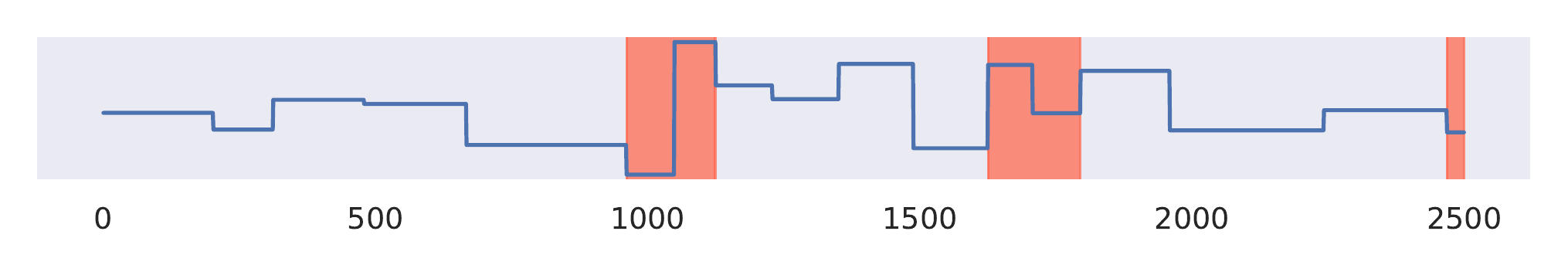}
\end{minipage}

\end{tabular}
\vspace{6mm}
\caption{Comparison of summaries created by two recent methods and our randomized method. The blue line shows the segment level importance scores with respect to time (frames). The orange areas indicate the frames selected for the final summary. All of the three methods use the same segment boundaries by KTS.}
\label{fig:summary_all_similar}
\end{figure*}

Figure~\ref{fig:summary_all_similar} shows examples of video summaries by two recent methods [25, 28] and our randomized method in Section 4.
All these methods uses KTS segmentation method.
As described in 4.3, we can see that all methods result in similar output regardless of the importance scores.

\subsection*{Visualization of Importance Score Correlation}
\begin{figure*}[t!]
    \centering
    \includegraphics[clip,width=0.3\linewidth]{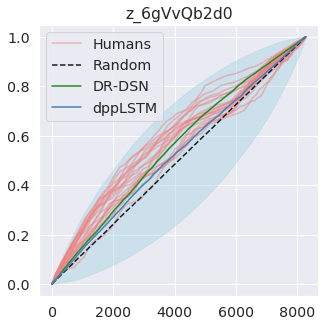}
    \includegraphics[clip,width=0.3\linewidth]{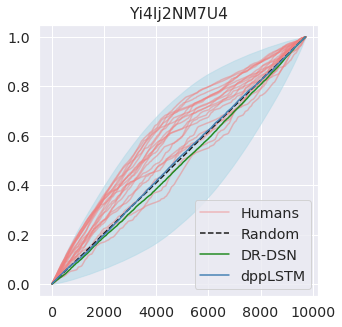}
    \includegraphics[clip,width=0.3\linewidth]{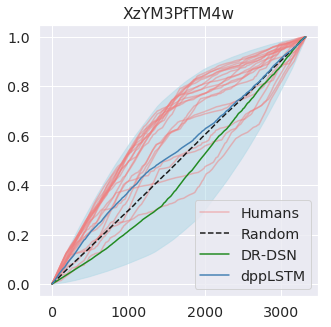}
    \\
    \includegraphics[clip,width=0.3\linewidth]{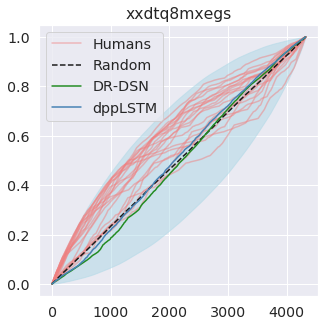}
    \includegraphics[clip,width=0.3\linewidth]{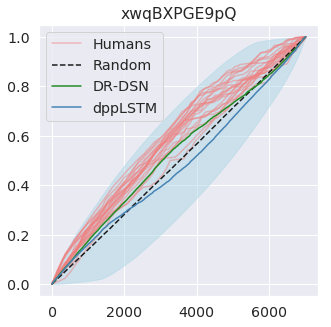}
    \includegraphics[clip,width=0.3\linewidth]{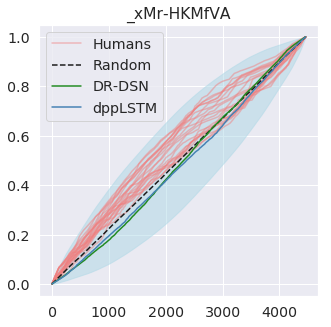}
    \\
    \includegraphics[clip,width=0.3\linewidth]{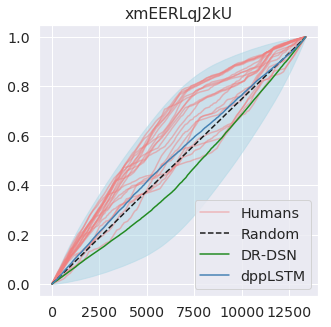}
    \includegraphics[clip,width=0.3\linewidth]{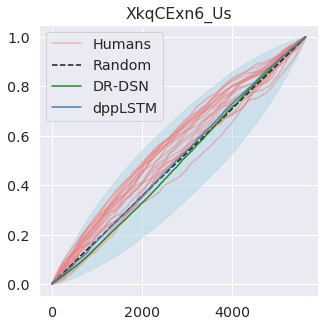}
    \includegraphics[clip,width=0.3\linewidth]{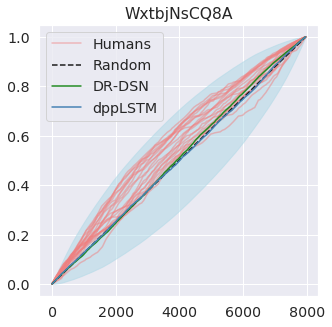}
    \\
    \includegraphics[clip,width=0.3\linewidth]{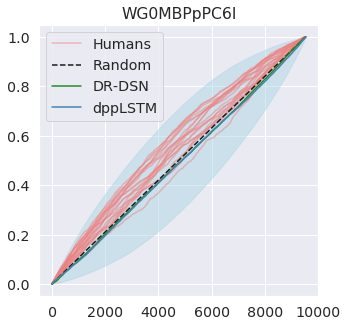}
    \includegraphics[clip,width=0.3\linewidth]{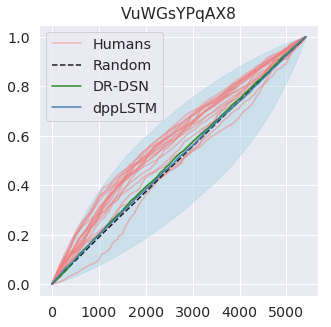}
    \includegraphics[clip,width=0.3\linewidth]{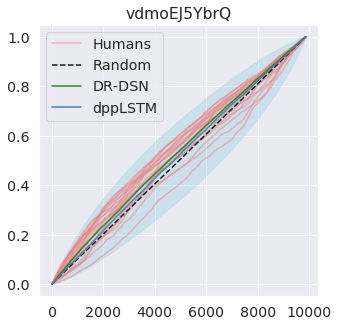}
    \caption{Example correlation curves. The red lines represent correlation curves for each human annotator and the black dashed line is the expectation for a random importance scores. The blue and green curves show the corresponding results to dppLSTM and DR-DSN methods, respectively.}
    \label{fig:cc}
\end{figure*}

Figure~\ref{fig:cc} shows examples of correlation curves on TVSum dataset.
The red lines show the $a_t$ curve for each human annotator and the black dashed line is the expectation for a random importance scores. The blue and green curves show the corresponding results to dppLSTM and DR-DSN methods, respectively.
Most examples show that human annotations, \ie, red lines, have positive correlation.

\begin{figure*}[t!]
    \centering
    \includegraphics[clip, width=0.8\linewidth]{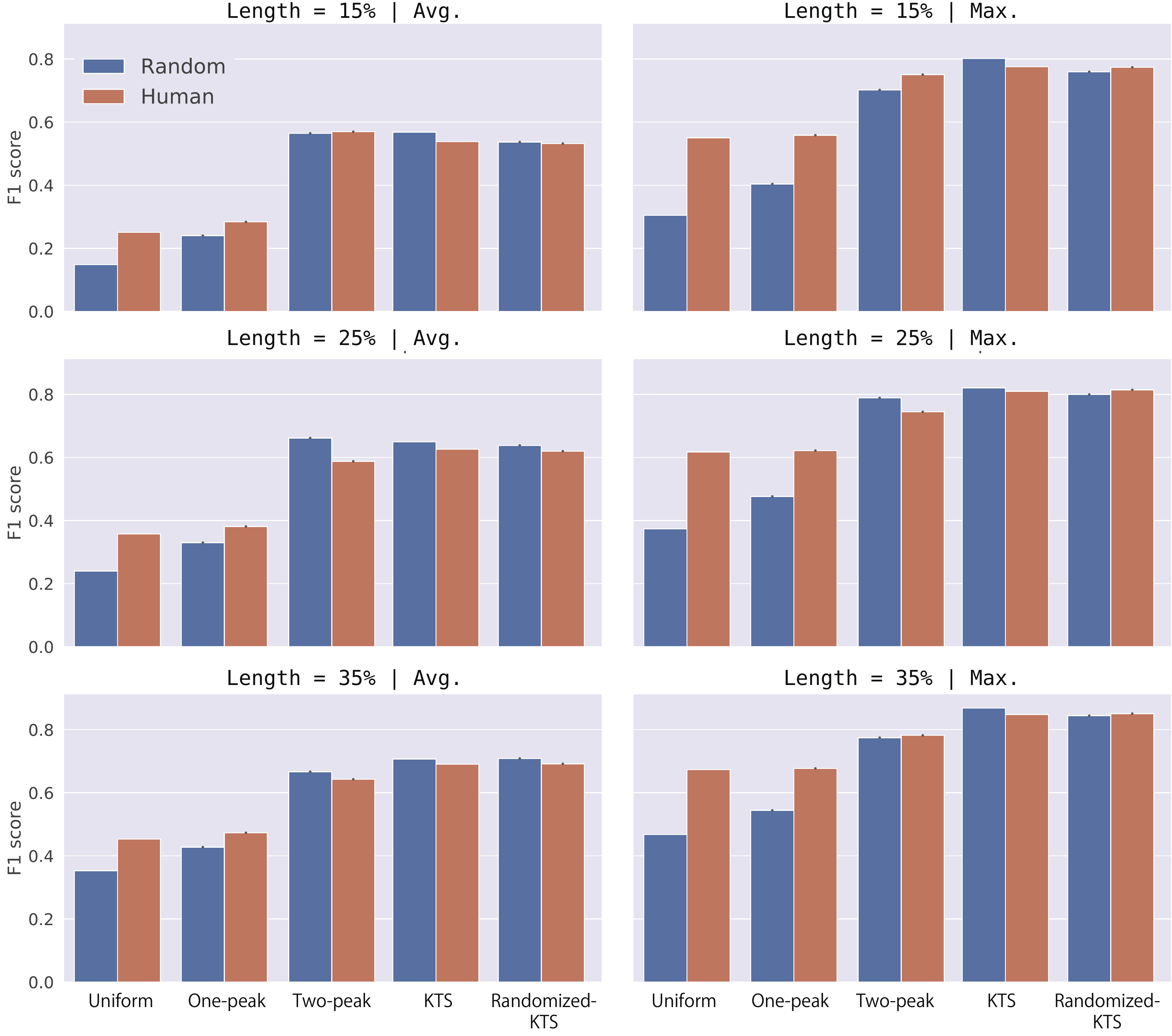}
    \caption{F1 scores for different segmentation methods combined to either random or human annotated importance scores (leave-one-out) for TVSum dataset. Blue bars refer to random scores and orange bars indicates human annotations. The results are computed with different summary length constraints, which are 15\%, 25\% and 35\% of original video length.}
    \label{fig:different_length}
\end{figure*}

\subsection*{Results with Different Summary Length}
Figure~\ref{fig:different_length} shows F1 scores of summaries generated with different summary length constraint; 15\%, 25\%, and 35\% of original video length.
The results are obtained by generating 10 summaries for each method.
For all summary length constraint, the results show similar trends. 
The randomized summaries have similar performance to human annotations for two-peak, KTS, and randomized KTS segmentation methods.
We also observed that F1 scores tend to get higher as summary length gets longer.

\subsection*{Human Evaluation}
\begin{figure*}[ht!]
    \centering
    \includegraphics[clip,width=\linewidth]{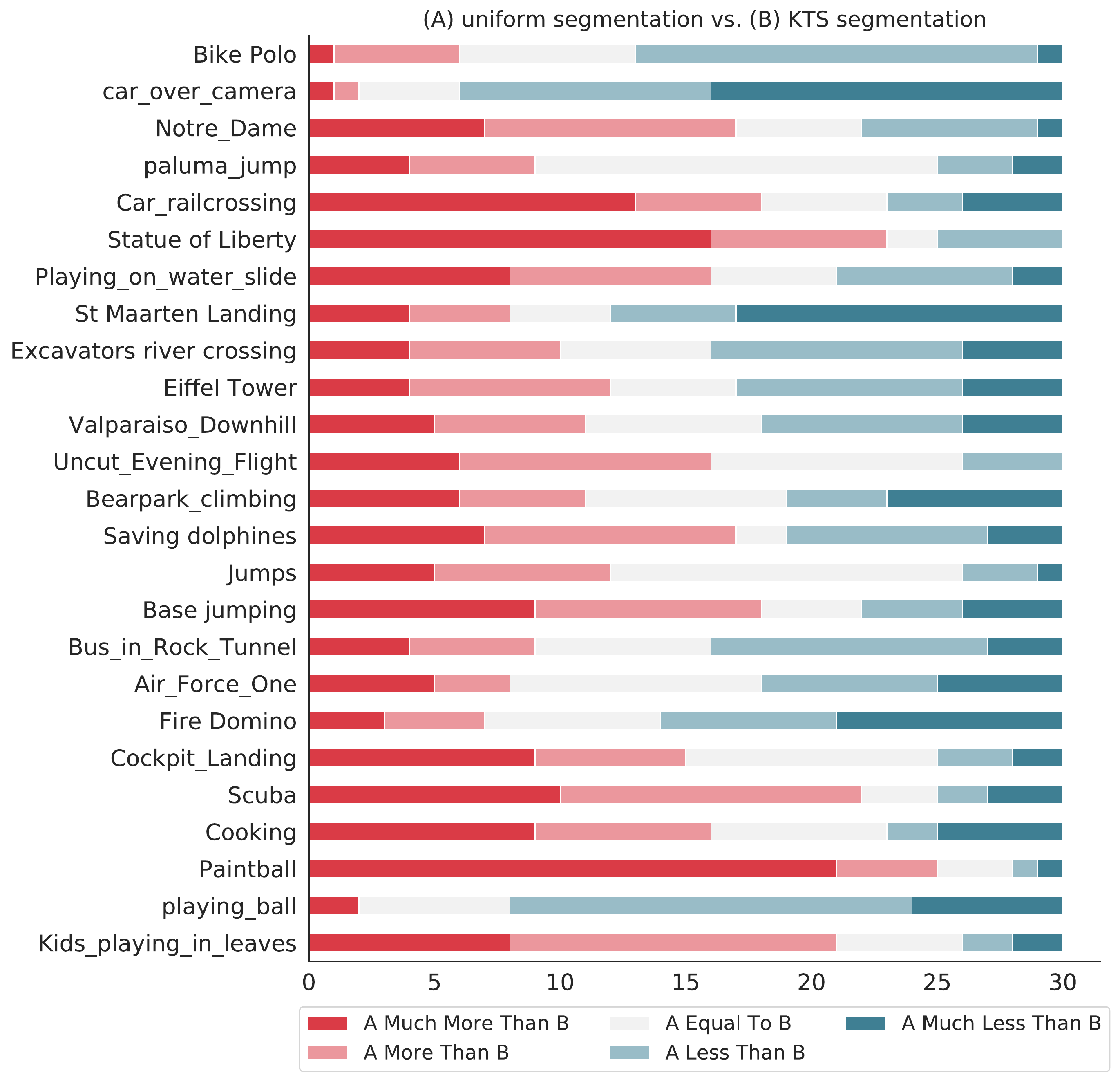}
    \caption{Comaprison of video summaries generated with A) uniform and B) KTS segmentation with random scores.}
    \label{fig:segment_comp_human_eval}
\end{figure*}

In the human evaluation, subjects watch two video summaries generated by two different methods and are asked, \textit{Which video better summarizes the original video subjects?}
We employed 30 subjects for each video summary pair on a crowdsourcing service.
Subjects selected their answer from -2 (A is much more than B) to 2 (A is much less than B).
Therefore, the averaged score larger than 0 indicates that method A is better otherwise, subjects prefer B better.
The results comparing KTS and uniform segmentation with random importance scoring are show in Figure~\ref{fig:segment_comp_human_eval}.
Overall, the averaged score is 0.19, thus subjects prefer video summaries using uniform segmentation.
In particular, subjects prefer uniform segmentation for videos recording long activity, \eg, sightseeing of the statue of liberty and scuba diving.
On the other hand, KTS works better for videos with notable events or activities.
For such videos, the important parts have little ambiguity, therefore the F1 scores based on the agreement of selected frames can get higher.

\begin{figure*}[ht!]
    \centering
    \includegraphics[clip,width=\linewidth]{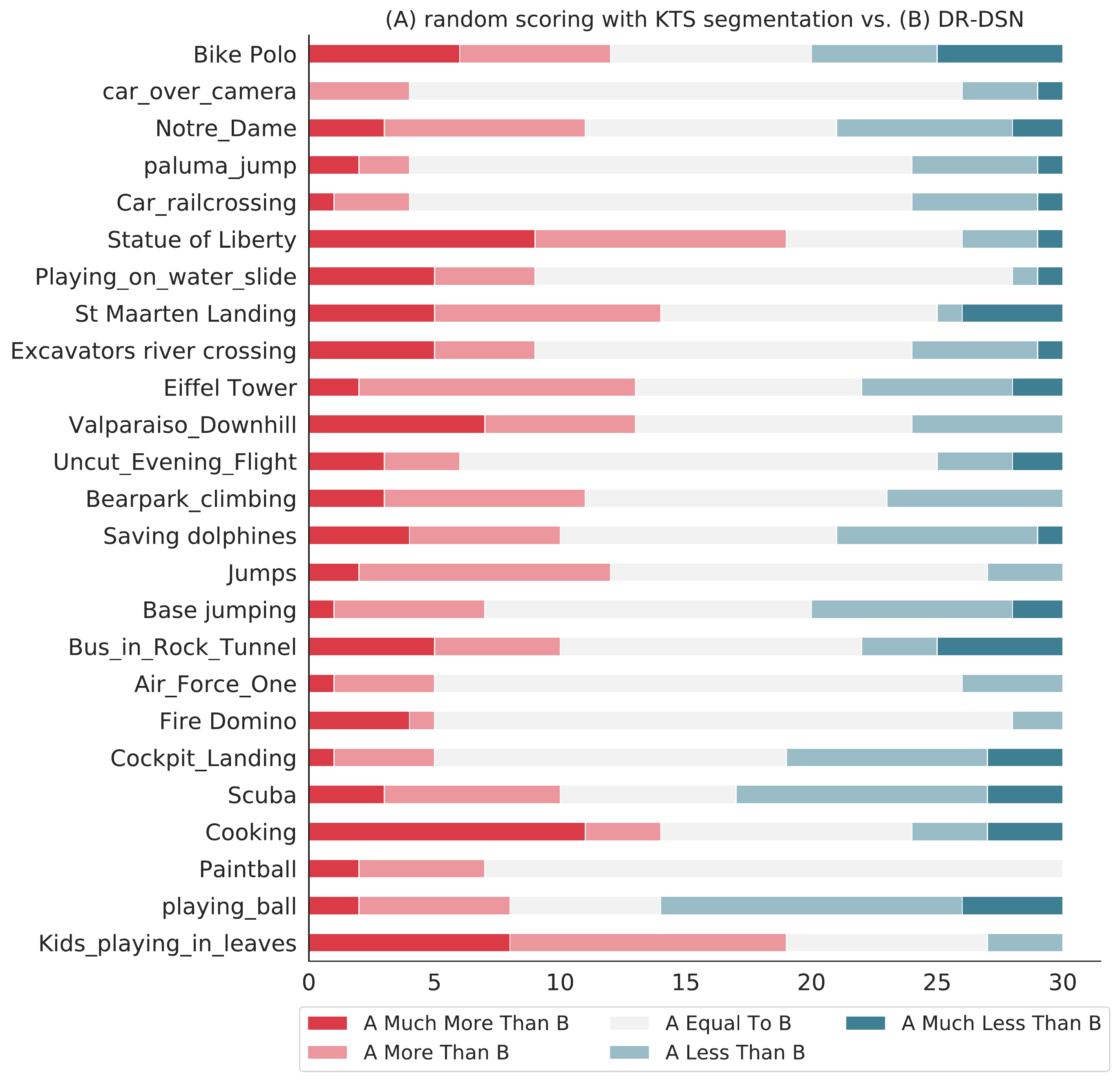}
    \caption{Comparison of video summaries generated with (A) random and (B) DR-DSN importance scores.}
    \label{fig:scoring_comp_human_eval}
\end{figure*}

Figure~\ref{fig:scoring_comp_human_eval} shows the results of comparing video summaries generated with random and DR-DSN scoring.
Both methods use KTS segmentation.
Random approach obtained a slightly higher score than random DR-DSN, however, 46\% of answers were that the summaries are equally good (bad).
This result further supports our findings that the importance scoring hardly affects the performance with a certain approach.

\begin{figure*}[ht!]
    \centering
    \includegraphics[clip,width=\linewidth]{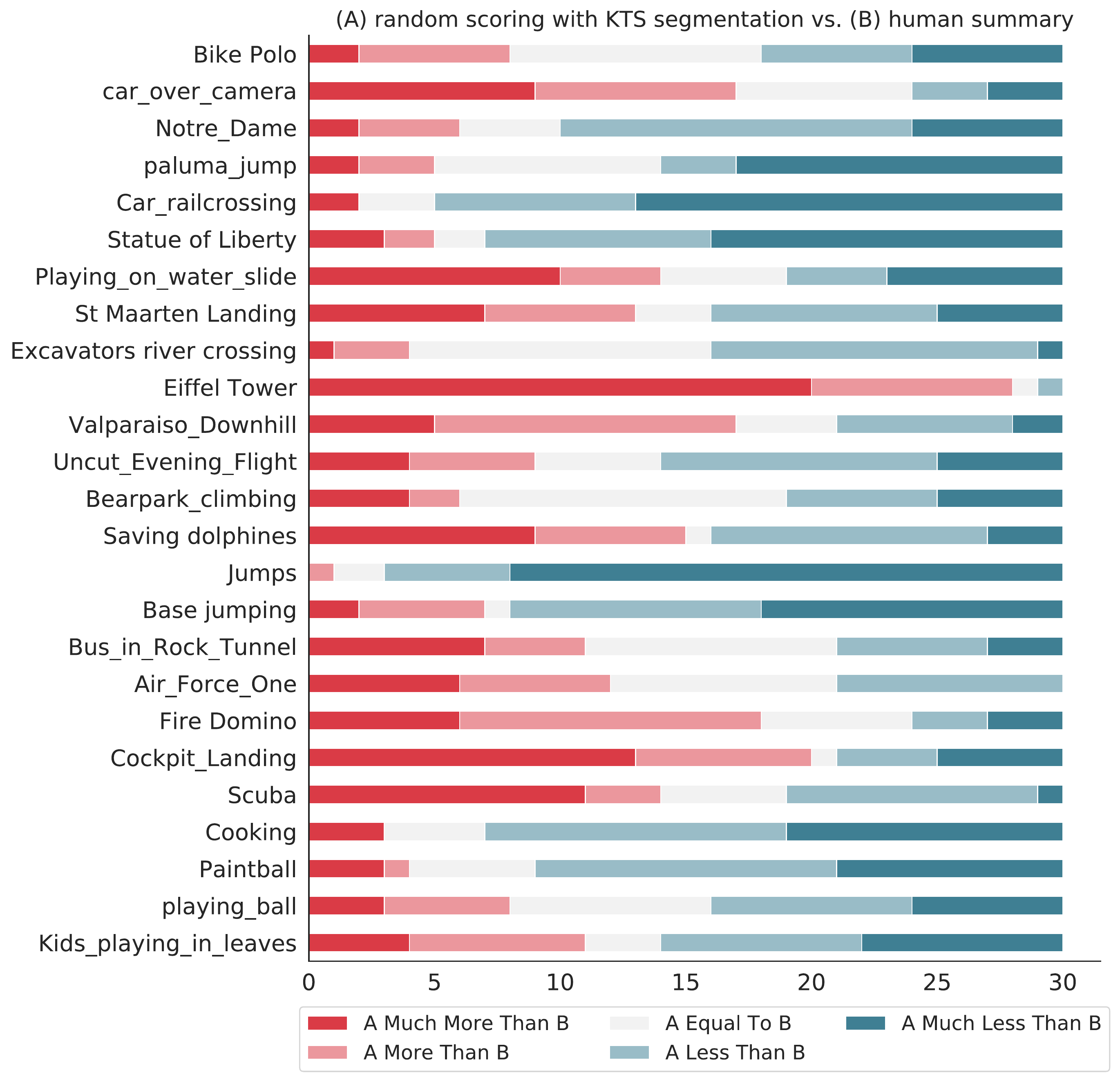}
    \caption{Comparison of video summaries generated with (A) random scoring with KTS segmentation and (B) human summaries.}
    \label{fig:random_vs_human_eval}
\end{figure*}

Other results of comparing video summaries generated with random scoring and manually created summaries are shown in Figure~\ref{fig:random_vs_human_eval}.
Overall, the averaged score is -0.17, thus subjects slightly prefer human summaries.
However, it is important to note that the answers are diverse among subjects; most videos got both ``A Much More Than B'' and ``A Much Less Than B.''
The results demonstrate the challenges in video summary evaluation.
The quality of video summaries often depends on subjects.
We also collected subject's free-form text feedback.
We observed that some subjects focus on the coverage of events in the original video, and some value inclusion of many scenes with main objects or people.
Evaluating a video summary from several aspects mentioned in the subjects' comments should be essential.

\subsubsection*{Rank Order Statistics and Human Evaluation}
One criticism for using rank order statistics for evaluating importance scores is that the metric might be irrelevant to the final quality of a video summary.
We compute Spearman's $\rho$ for DR-DSN on the SumMe dataset.
For reference importance scores, we employ the ratio of human annotators who selected the frame for their manually-created summary.
We split the video into two groups; one is for videos where DR-DSN shows a positive correlation to reference scores, and otherwise.
For each group, we investigate the performance scores against randomized summaries.
Videos with positive correlation are better rated than those with negative correlation.
The average score of the group with positive correlation is 0.13, and the other is 0.22.
In this result, a lower score means that video summaries generated by DR-DSN are better against ones with random scoring.
This observation suggests that the higher correlation of importance scores can result in better quality of final output.

\end{document}